\title{\bf Evolving Spiking Networks with Variable Resistive Memories}

\documentclass{article}
\usepackage{ecj,epsfig,latexsym,natbib,epstopdf,subfig}

\DeclareGraphicsExtensions{.eps,.png,.EPS,.PNG}
\begin{document}

\title{Evolving Spiking Networks with Variable Resistive Memories}
\begin{center}
\author{
Gerard Howard, Larry Bull, Ben de Lacy Costello, \\Andrew Adamatzky, Ella Gale \\
University of the West of England, \\
Bristol BS161QY UK \\}
\end{center}
\date{}
\maketitle

\begin{abstract}

Neuromorphic computing is a brainlike information processing paradigm that requires adaptive learning mechanisms.  A spiking neuro-evolutionary system is used for this purpose; plastic resistive memories are implemented as synapses in spiking neural networks.  The evolutionary design process exploits parameter self-adaptation  and allows the topology and synaptic weights to be evolved for each network in an autonomous manner.  Variable resistive memories are the focus of this research; each synapse has its own conductance profile which modifies the plastic behaviour of the device and may be altered during evolution.  These variable resistive networks are evaluated on a noisy robotic dynamic-reward scenario against two static resistive memories and a system containing standard connections only.   Results indicate that the extra behavioural degrees of freedom available to the networks incorporating variable resistive memories enable them to outperform the comparative synapse types.\\
\end{abstract}

\section{Introduction}

Neuromorphic Computing (NC)~\citep{meadneuromorphic}  is a bio-inspired paradigm concerned with emulating brainlike functionality within artificial systems.  Typical NC involves the use of a physical network comprised of neurons (e.g. CMOS~\citep{cmos-overview}) that are interconnected by a dense web of nanoscale synapses.  Resistive Memories (RMs)~\citep{ionics-RSM} are synapse-candidate devices --- typically Metal-Insulator-Metal ---  that can be induced to switch to one of several resistances through application of an appropriate voltage.  They can be manufactured at the nanoscale and provide a nonvolatile memory whereby the state (resistance) of the device can vary depending on its activity.  Nonvolatile memory faciliates low-heat, low-power storage~\citep{nonvol-mem-mem}, alleviating typical nanoscale concerns such as power usage and heat dissapation. A context-sensitive dynamic internal state allows synapse-like information processing.  We categorise RMs as either Resistive Switching Memories (RSMs)~\citep{ionics-RSM} or memristors (memory-resistors)~\citep{chua-mem} depending upon their characteristic behaviour.  RSMs allow switching between (usually two) discrete resistance states, whereas memristors permit gradual traversal of a nonlinear resistance profile.

NC requires some form of in-trial learning to harness the computational power of the network --- typically a form of Hebbian learning~\citep{hebb43} is used to realise Spike Time Dependent Plasticity (STDP)~\citep{stdp}.  RM synapses alter their efficacy during the lifetime of the network, depending on the activity of the neurons they are connected to.  The Hebbian mechanism is coupled with a neuro-evolutionary model that allows network topologies to be modified during the application of a Genetic Algorithm (GA)~\citep{holland75ga}.  Self-adaptive search parameters are shown to provide a flexible learning architecture which may be especially beneficial given the autonomous nature of NC.

A {\em variable RM}, the focus of this study, is an RM whose STDP response can be tuned by evolution,  potentially imparting a variety of adaptive behaviours to the networks.  Previous studies~\citep{howardTEC} have indicated that fixed STDP profiles can be exploited by evolution and cast into specific roles, such as facilitating or depressing synapses, with synaptic role based on STDP response ---  it follows that more varied responses may permit more finely-tuned behaviours within the networks.

The computational properties of two types of variable RM --- memristor and RSM  ---  are analysed when cast as synapses in evolutionary Spiking Neural Networks (SNNs~\citep{spiking-n-m}).  Our hypothesis is that the additional degrees of functional freedom afforded to the variable RM networks can be harnessed by the evolutionary process.  To test this hypothesis, the variable memristor and variable RSM networks are compared to networks comprised of (i) PEO-PANI memristors~\citep{peo-0}, (ii) HP memristors,~\citep{missing-mem-found}, and (iii) constant, non-plastic connections.  A dynamic simulated robotics navigation task is selected for this purpose.  To our knowledge, this is the first approach that allows for the self-adaptation of the characteristic performance of the RMs alongside neuroevolution of both neurons and connection structure.

\section{Background}
\label{background}

\subsection{Spiking Networks}
\label{snn+esnn}
Spiking Neural Networks (SNNs) present a phenomenological model of neural activity in the brain.  In an SNN, neurons are linked via unidirectional, weighted connections.  Each neuron has a measure of excitation, or {\em membrane potential} and communicates via the voltage spike, or {\em action potential}.  A neuron spikes when its membrane potential exceeds some threshold, which typically requires a cluster of incoming spikes arriving within a short time period.  A spike emitted from a neuron is received by all connected postsynaptic neurons.  

As the membrane potential may be considered a form of memory, such networks are able to produce temporally dynamic activation patterns which potentially allows for increased computing power~\citep{maass,sag-wex} when considering temporal problems (e.g. robotics, time series analysis), compared to ``stateless'' network models such as the Multi Layer Perceptron~\citep{rumelhart-mcclelland} (although Continuous Time Recurrent Neural Networks can use internal dynamics to the same effect e.g.~\citep{tmaze}).  SNNs are preferred because the voltage spike is an efficient medium of  communication when compared to traditional schemes where voltage is constantly applied to a connection.  The benefits of such a scheme --- low-heat, low-power communication --- are heightened when coupled with nonvolatile synapses (such as the Resistive Memories used herein), as sparse pulse-based encoding schemes may be envisioned.

Two well-known formal SNN implementations are the Leaky Integrate and Fire (LIF) model and the Spike Response Model (SRM)~\citep{spiking-n-m}.  Neuro-evolution applies evolutionary techniques to alter the topology/weights of neural networks.~\cite{neuroevo-arch-learn} survey various methods for evolving both weights and architectures.~\cite{evo-rob-book} describe the evolution of networks for robotics tasks.

\subsection{Resistive Memories}
\label{RSM}

Numerous Resistive Memories have been previously manufactured from a plethora of materials ---~\cite{akinaga-reram} provide a summary.  Resistive Switching Memories are predominantly metal oxides (HfO$_{2}$, Cu$_{2}$O, ZnO, ZrO$_{2}$, TiO$_{2}$).  Memristor materials are more varied and include conductive polymers~\citep{peo-0},  metal silicides~\citep{nano-mem-syn-neuromorphic}, and crystalline oxides~\citep{macro-memristor} in addition to certain metal oxides. 

\begin{figure}[t!]
\begin{center}
\subfloat[]{ \epsfig{file=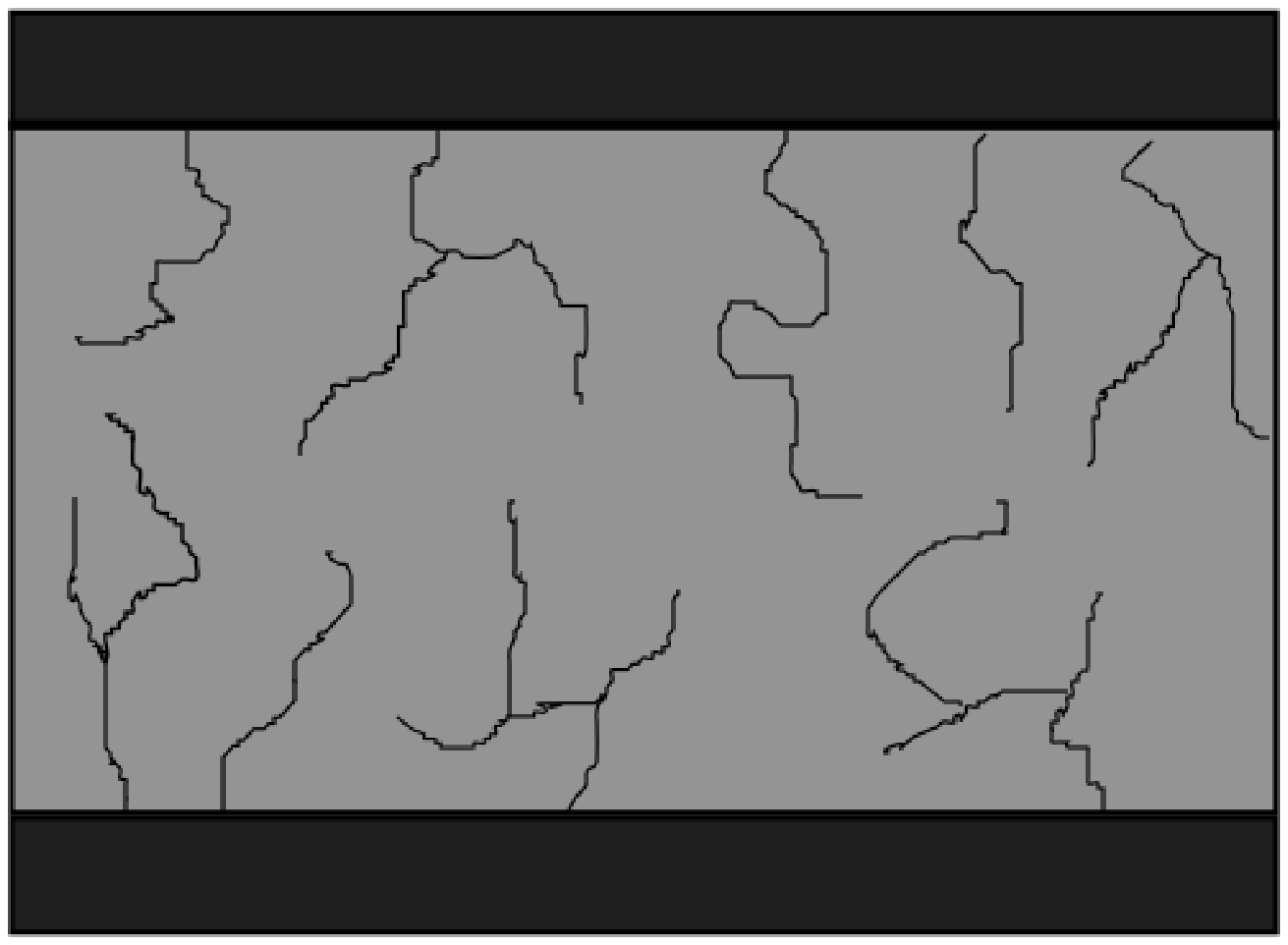,width=4.45cm, height=4.45cm}}
\subfloat[]{ \epsfig{file=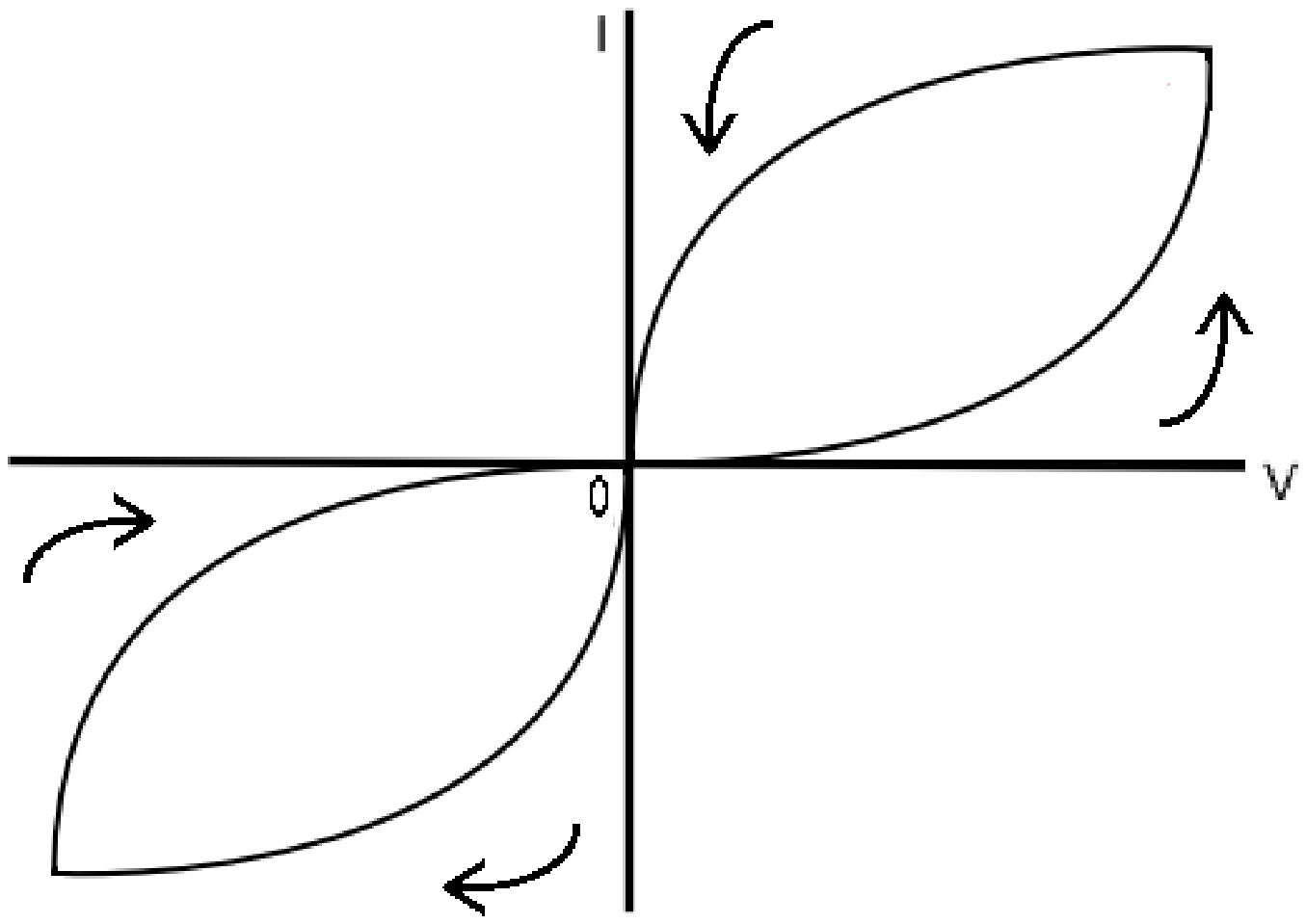,width=4.45cm ,height=4.45cm}}\\
\subfloat[]{ \epsfig{file=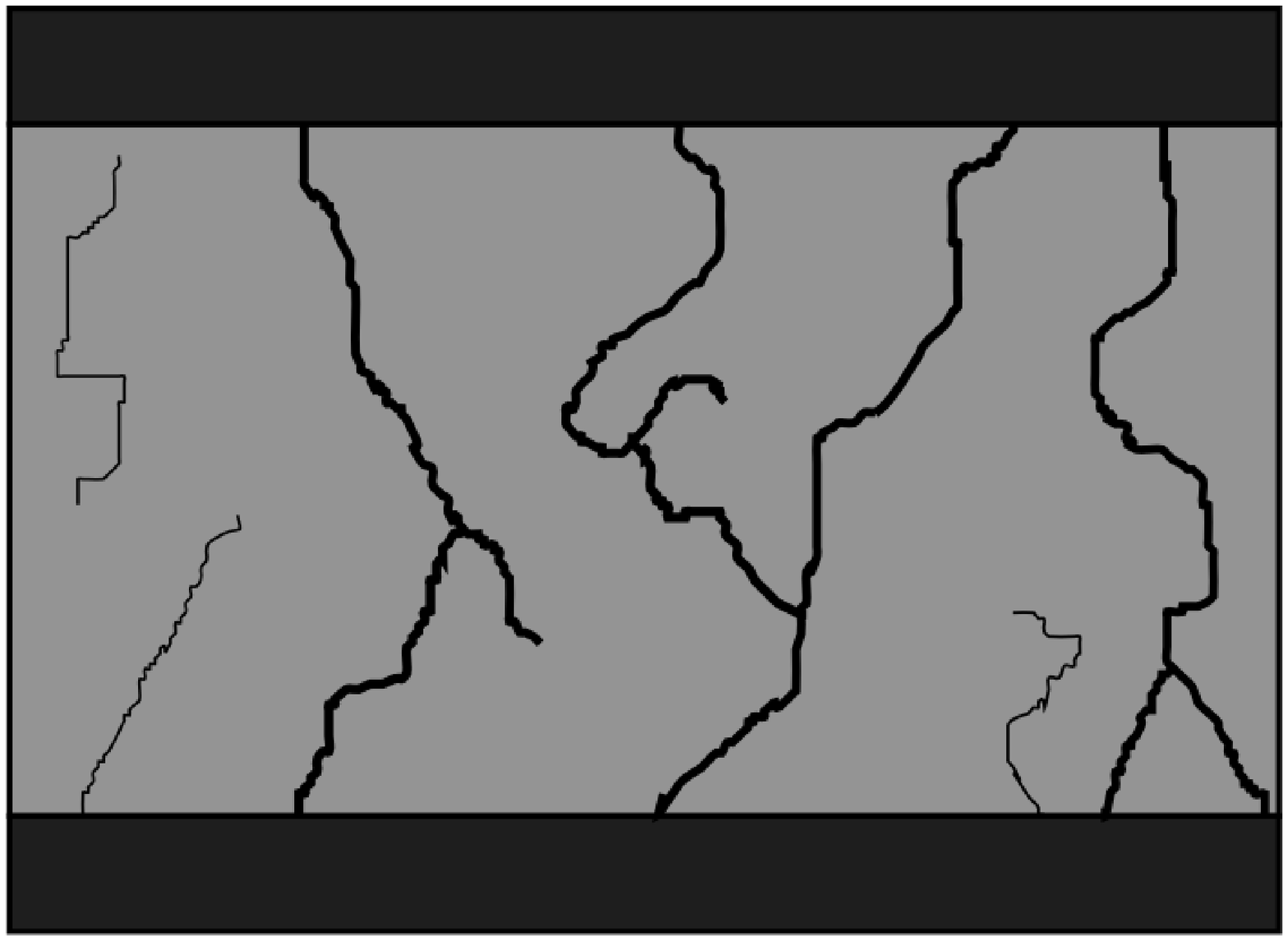, height=4.45cm, width=4.45cm}}
\subfloat[]{ \epsfig{file=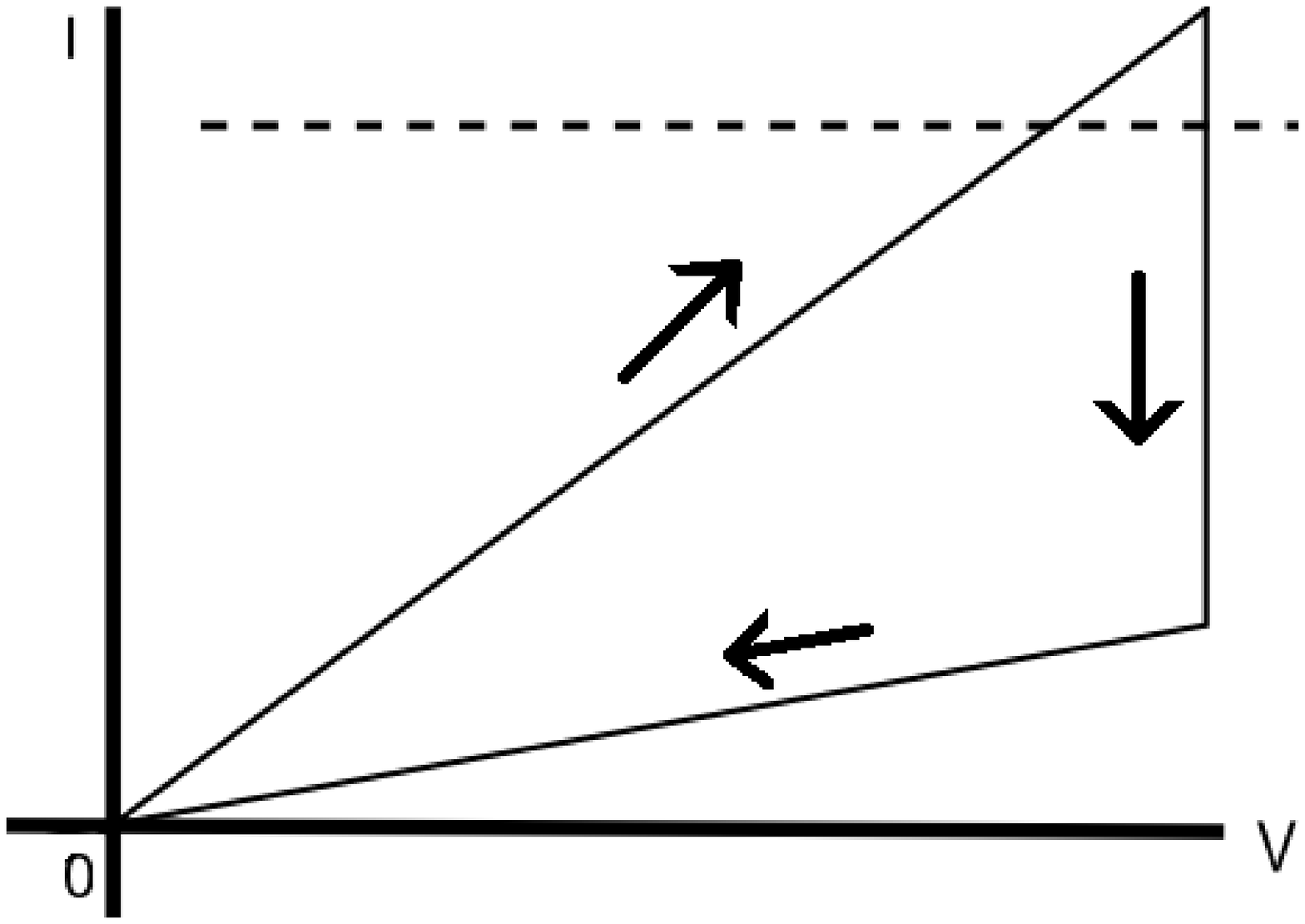,width=4.45cm, height=4.3cm}}\\
\end{center}
\caption[]{(a)  Incomplete filament formation (memristor) --- as there are no complete filaments, ionic conductivity gives rise to (b)  Generalised HP memristor I-V curve(c) Complete filament formation (RSM); dark lines show complete filaments (d)  Generalised RSM I-V curve, dashed line showing a possible current compliance.  The steeper gradient is the LRS, the shallower is the HRS.}
\label{rsm-compare}
\end{figure}

One popular theory, espoused by~\cite{ionics-RSM}, states that the resistance profiles of both types of RM are considered to be the result of the appearance of filaments in the substrate, which may arise due to material defects or conditions during synthesis.  Filaments are conductive pathways through the material that allow electrons to flow through them.  In our taxonomy, memristors (Fig.~\ref{rsm-compare}(a)) do not form complete filaments, giving rise to the characteristic nonlinear I-V curves of these devices as other mechanisms (such as ionic conductivity) play a more prominent role in electron transport (Fig.~\ref{rsm-compare}(b)).  Complete filament formation occurs in the case of RSMs (Fig.~\ref{rsm-compare}(c)) which results in ohmic I-V profiles (Fig.~\ref{rsm-compare}(d)). This distinction is not universal (e.g. under specific conditions an RSM may act like memristor, and {\em vice versa}).

\begin{figure}[t!]
\begin{center}
 \epsfig{file=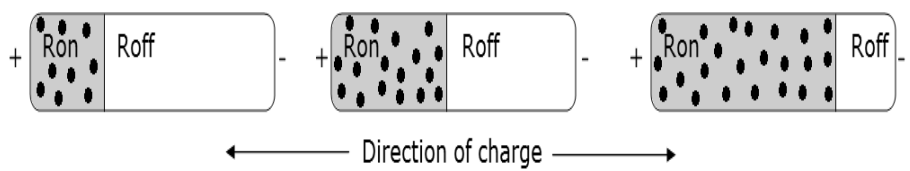,width=12cm, height=2.5cm}
\end{center}
\caption[]{Generalised memristor model shown as two resistors whose boundary, and hence final resistance, varies depending on the polarity of the input charge.  Resistor  $R_{on}$ contains oxygen vacancies (black dots) which act as charge carriers and possesses low resistance, resistor $R_{off}$ has no vacancies and a high resistance.  From an initial boundary position (Centre), charge flowing from the positive terminal to negative terminal causes the vacancies to percolate towards the negative terminal (Right), increasing the size of $R_{on}$ which decreases the resistance of the device.  The opposite is true for charge flowing from the negative to the positive port. (Left)}
\label{hpmem-vacancies}
\end{figure}

\subsubsection{Memristors}
\label{memristors}

The memristor, a class of RM theoretically characterized by~\cite{chua-mem}, has recently enjoyed a resurgence of interest after being manufactured from titanium oxide by HP labs~\citep{missing-mem-found}.  Memristors are the fourth fundamental circuit element, joining the capacitor, inductor and resistor.  Like RSMs, memristors can be operated as binary switches; we forgo this method of operation and use the memristor as an analogue device capable of incremental resistance alteration. 

As filaments are not formed, some other conductivity mechanism is used instead; in this case ionic conductivity gives rise to the nonlinear I-V curve seen in Fig.~\ref{rsm-compare}(d).  An explanatory model of the titanium dioxide device can be seen in Fig.~\ref{hpmem-vacancies}.  Here, the memristor is modeled as two variable resistors  $R_{{\mathrm on}}$ and  $R_{{\mathrm off}}$. The instantaneous resistance of the device can be attributed to the position of the boundary between $R_{{\mathrm on}}$ (which is doped with oxygen vacancies (charge carriers) and therefore has low resistance) and $R_{{\mathrm off}}$ (which displays higher resistance).  Charge flowing through the memristor in a given direction causes the oxygen vacancies to migrate in that direction, moving the boundary between $R_{{\mathrm on}}$ and $R_{{\mathrm off}}$ and altering the resistance of the device.  Nonvolatility arises due to the chemical nature of the mechanism;~\cite{missing-mem-found} provide more details

Formally, a memristor is a passive two-terminal electronic device that is described by the relation between the device terminal voltage $v$, terminal current $i$ (which is related to the charge $q$ transferred onto the device), and magnetic flux $\varphi$, as (1) shows.  Resistance can be made to increase or decrease depending on the polarity of the voltage.  The nonvolatile resistance, $M$, is a nonlinear function of the charge (2).  \\

$v = M(q)i$                      \hfill (1)\\

$M(q) = d\varphi(q) / dq $   \hfill(2)\\

Previous applications of memristors within neural paradigms are ubiquitous: Titanium dioxide memristors~\citep{missing-mem-found} have been used in the manufacture of nanoscale neural crossbars~\citep{comp-resis-xbar}, and silver silicide memristors have been shown to function in neural architectures~\citep{nano-mem-syn-neuromorphic}.  Other successful applications include the modelling of learning in amoeba~\citep{mem-amoeba}, as well as  pattern recognition by crossbar circuits for robotic control~\citep{mem-pattern-recognition}.  In particular,~\cite{mem-pattern-recognition} highlights the attractive prospect of applying evolutionary computation techniques directly to memristive hardware, as memristors can simultaneously perform the functions of both processor and memory.    

\subsubsection{Resistive Switching Memories}
\label{unipolar}

As their nomenclature implies, RSMs are typically used as a bistable resistive switch; pulsing with a voltage over some threshold transfers the device from an initial Low-Resistance State (LRS) to a High-Resistance State (HRS).  Successive voltage pulses can then be used to switch between these states; the amount of voltage required depends on the physical properties of the device.  RSMs are typically Metal-Insulator-Metal devices that comprise an ion-conducting semiconductor, sandwiched between two layers of metal.  Electroforming creates complete filaments which provide ohmic conductance in the LRS.  Driving over a threshold voltage breaks these filaments and transfers the device to the HRS.  A further voltage spike (usually in conjuction with a current compliance to protect the device) reforms these filaments and reinstates the LRS --- see Fig.~\ref{rsm-compare}(c)-(d).  RSMs follow (1) and (2), under the proviso that $M$ is now a linear function of charge.  RSMs are not sensitive to the polarity of voltage used.  Examples of RSMs are confined to binary operation, e.g. as Resistive Random Access Memory~\citep{akinaga-reram, hosoi-reram}, although neural implementations exist~\citep{guangpu}.  

\subsection{Synaptic Plasticity}
\label{plasticity}

Hebbian learning ~\citep{hebb43} is thought to account for adaptation and learning in the brain.  Briefly, Hebbian learning states that ``neurons that fire together, wire together'' --- in other words in the event that a presynaptic neuron causes a postsynaptic neuron to fire, the synaptic strength between those two neurons increases so that such an event is more likely to happen in the future.  Such a mechanism allows for self-organising, correlated activation patterns.

Spike Time dependent Plasticity (STDP)~\citep{stdp} was originally formulated as a way of implementing Hebbian learning within computer-based neural networks.  Interestingly, the STDP equation has been found to have distinct similararities to the reality of Hebbian learning in biological synapses~\citep{bi-poo}.  It has recently been postulated that a memristance-like mechanism affects synaptic efficacy in biological neural networks~\citep{linares-barranco}, based on similarities between memristive equations and their neural counterparts.~\cite{querlioz} have recently shown that memristive STDP can be made to mitigate device variations that are currently intrinsic to hardware memristor realisation. 

Typically when implementing STDP with Resistive Memories of both kinds, a bidirectional voltage spike is emitted by a neuron whose membrane potential exceeds some threshold.  Bidirectional spikes are a necessity as they allow the temporal coindicence of spikes across a synapse to be tracked.  The spike can be approximated by either a continuous~\citep{stdp-nano-cmos-asyn,linares-barranco} or discrete~\citep{nano-mem-syn-neuromorphic,stdp-discrete} waveform through time, which is transmitted to all synapses that the neuron is connected to (presynaptic or postsynaptic).  In the case of memristors, if the instantaneous voltage across a synapse surpasses some threshold --- typically when the waveforms sufficiently overlap --- the conductance of the synapse changes.  For RSMs, multiple consecutive voltage spikes of a given polarity within a short time frame are required to switch the device from one state to the other.  Note that this removes the element of biological realism from RSM STDP whilst providing a fast-switching binary behaviour.  

The notion that varied plastic behaviours could be combined in a single network is an attractive one from a computing perspective, as more functional degrees of freedom may be afforded to the synapses.  Additionally, certain synaptic behaviours may be more beneficial than others in certain positions within the network.  Integration of neuroevolution with heterogeneous neuromoduation rules is investigated by~\cite{soltoggio08}, and has been extended to robot controllers~\citep{neuromod-robot}.   Increased behavioural diversity (and high-quality pathfinding in the latter case) is reported.  Probabilistic spike emission which is governed by modulatory Hebbian rules has also been investigated~\citep{maass-zador}.  The authors show a biologically-plausible mechanism capable of computing with short spike trains where the population of synapses display heterogeneous probabilities of transmitting/blocking a spike.~\cite{flor-urz} present a nodes-only encoding scheme where synapses are affected by four versions of the Hebb rule, which generates pathfinding behaviour online from initially random actions.  Synaptic weights are not evolved; instead evolution is performed on the rules which govern how synapses react to STDP.  High adaptability to new environments is evidenced.  Since synaptic strength is not directly modelled and all synapses at a given node display homogenous STDP behaviour, it is unclear how to transition such a scheme to memristive/hardware implementations.   ~\cite{howardTEC} used memristive STDP to vary the behaviour of a synapse (and therefore the network) during a trial.  In this study we extend this concept to allow for variable RMs of both types, whereby the STDP response of the synapse can be tailored by evolution to suit its role within the network.  

\subsection{Viability of a Variable Resistive Memory}
Conceptually, a variable RM can be seen as a nonvolatile, low power, and behaviourally diverse synapse candidate for neuromorphic hardware.  A lingering question is then: ``How do we know that variable Resistive Memories are physically realisable?''.  Although the field is very much in its infancy, there are a number of reasons to believe that the networks produced will have a physical analog.

The STDP response of the memristors can be encapsulated in a ``physical properties'' parameter, where changing this parameter varies the behaviour of the synapse.  The memristor ``physical properties'' parameter $\beta$ (first used by~\cite{howardTEC}, derived from the original equations by~\cite{missing-mem-found}) is given as $\gamma^{v}/D^2$, where $\gamma^{v}$ is the mobility of oxygen vacancies and $D$ is the device thickness in nanometres.~\cite{gale2012a} presents a more detailed model which indicates that varying the electrode size also affects memristance, and has some experimental verification~\citep{gale2012b}.  Due to the relationship between $\beta$ and $D$, it can be inferred that a smaller $D$ will generate a larger memristive effect: as memristiance is more pronounced at smaller scales, many memristive devices are therefore likely to be manufactured at compatible (nano) scales.  As $\beta$ was originally derived from modelling of physical devices,  $\beta$ values in the range of those proposed in this article are potentially viable.  Variations of a small $D$ can also create a greater potential variance in $\beta$ for a given $\gamma^{v}$, meaning that  the amount of synaptic variance available to any newly-found memristor is predicted to increase with increasing miniturisation.  Finally, the sizes of $D$ given for current memristor models predict a scale that is small enough to permit sufficient synaptic density for neuromorphic hardware implementations --- equivalent to that found in the human brain. 

Another possible method of varying the STDP response of a memristor requires irradiation by an ion beam, as described by~\cite{varmem-nuke}.  Simulations of irradiated titanium dioxide memristors were found to possess a lower oxygen vacancy mobility ($\gamma^{v}$) and reduced resistance in the doped region ($R_{{\mathrm{on}}}$).  Additionally, selective bombardment of specific synapses could alter synaptic behaviour in an online manner, although the precise method of targetted radiation delivery would depend upon the physical structure of the network.  Advantages of this approach include online behaviour modification, the ability to elicit varied STDP responses from a homogenous group of memristors postfabrication, and potentially including this mechanism as part of a feedback loop during evolution.  

RSMs could be adapted in a similar manner to $\beta$, where varying the device would in this case change its switching sensitivity.  As RSMs can be made from titanium dioxide, they are also candidates for control via irradiation~\citep{varmem-nuke}.  A benefit of the ``physical properties'' parameter is that, as it is implicitly grounded in reality, the dimensions of the RM corresponding to a given parameter value can be calculated and used as a ``best fit'' for reconstructing simulated natworks with available hardware devices.  This is especially pertinent as our research group is capable of creating and profiling both types of RM~\citep{gale2012c}.

In summary, (i) RMs are suitable synapse candidates, (ii) STDP is a popular means to achieving learning within RM spiking networks, and (iii) motivation for researching a variable RM has been given. 

\section{The System}
\label{the-system}
The system consists of 100 SNNs which are evaluated on a robotics test problem, and altered via a steady-state GA.  Each experiment lasts for 1000 evolutionary {\em generations}; two new networks are created and {\em trialled} on the test problem per generation.  Each trial consists of 8000 {\em timesteps}, which begin with the reading of sensory information and calculation of action, and end with the agent performing that action.  Every timestep comprises 21 {\em steps} of SNN processing, at the end of which the action is calculated.  The state of the system was sampled every 20 generations and used to create the results.  Results were averaged over 30 experimental repeats.  

\subsection{Control Architecture}
\label{snn-imp}
Spiking network implementation is based on the LIF model.  Neurons are arranged into a three-layer (input, hidden, output) network without recurrency but with hidden-hidden neuron connectivity.  Each neuron has a membrane potential $y$$>$0 which slowly degrades over time, and can be modified either by an external current or by spikes received from presynaptic neurons.  As spikes are received by the neuron, the value of $y$ is increased in the case of an excitatory spike, or decreased if the spike is inhibitory.  If $y$ surpasses a positive threshold $\theta_y$ the neuron spikes and transmits an action potential to every neuron to which it is presynaptic, with strength relative to the synaptic weight between those two neurons.  The neuron then resets its membrane potential to prevent oversaturation; for a given neuron, $y$ at time $t$ is given in (3):\\

$y(t+1) = y(t) + (I+a-by(t))$ \hfill(3)\\

$If (y > \theta_y) y = c$ \hfill(4)\\

Equation (4) shows the reset formula.  Here, $y(t)$ is the membrane potential at time $t$, $I$ is the input current to the neuron, $a$ is a positive constant, $b$ is the degradation (leak) constant and $c$ is the reset membrane potential of the neuron.  A model of temporal delays is used so that, in the single hidden layer only, a spike sent to a neuron not immediately neighbouring the sending neuron is received $x$ steps after it is sent, where $x$ is the number of neurons between the sending neuron and receiving neuron.  Each output neuron has an activation window that records the number of spikes produced by that neuron at each timestep.

\begin{figure}[t!]
\begin{center}
\subfloat[]{ \epsfig{file=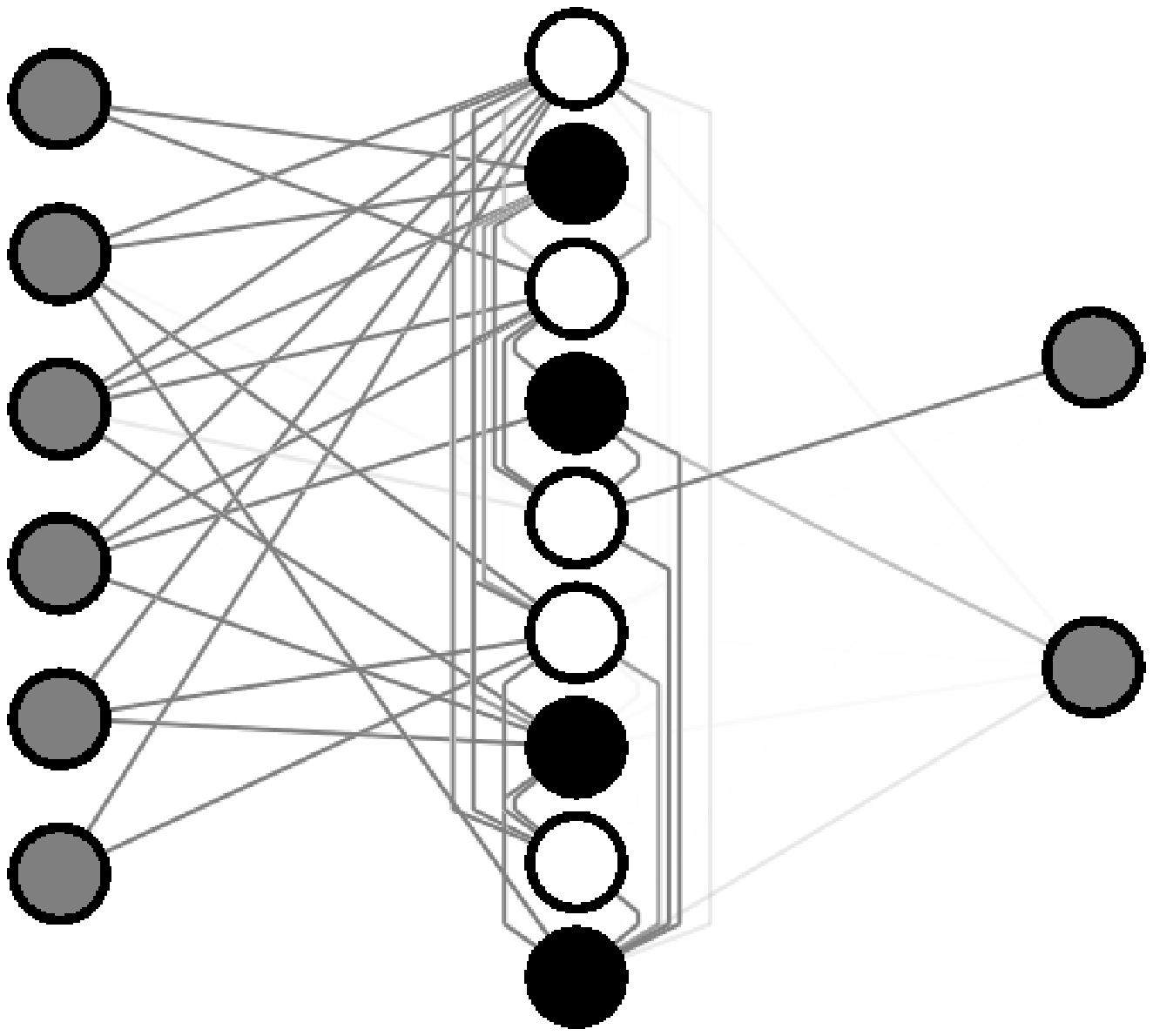,width=6cm,height=3.8cm}}
\subfloat[]{ \epsfig{file=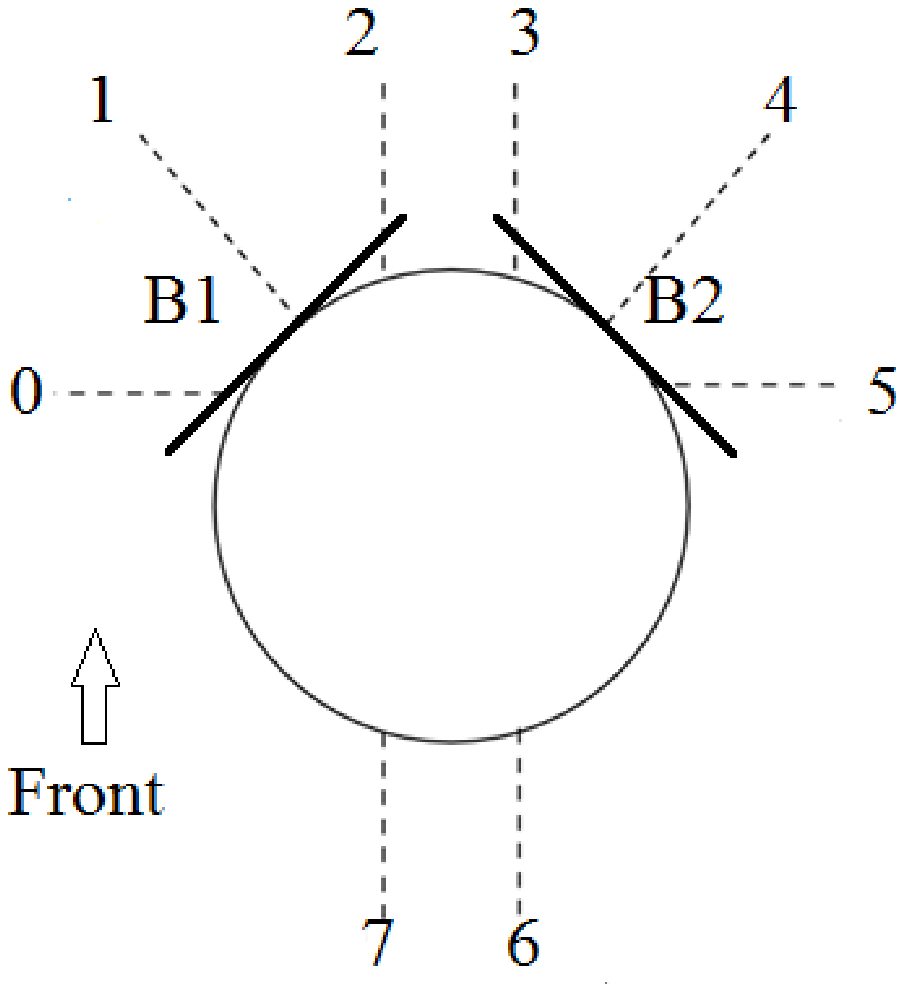, height=1.5in, width=1.5in}}
\end{center}
\caption[]{(a) A typical SNN architecture.  In the hidden layer, white neurons denote excitatory neurons and black neurons signify inhibitory neurons. (b) Khepera sensory arrangement.  Three light sensors and 3 IR sensors share positions 0, 2, and 5 and form the network input.  Two bump sensors, B1 and B2, are shown attached at 45 degree angles to the front-left and front-right of the robot.}
\label{khep}
\end{figure}

SNN parameters are $ initial\:hidden\:layer\:nodes=9$, $a=0.3$, $b=0.05$, $c=0.0$, $c_{ini}=0.5$, $\theta_y = 1.0$, $output\:window\:size=21$.  Previous experimentation has shown that these parameters, specifically the initial hidden-layer nodes and output window size, have relatively little effect on network performance.  Fewer initial hidden-layer neurons can restrict possible topologies (and therefore possible output responses) during early generations, resulting in a slightly longer learning process.  Output window size is experimentally set to allow a reasonable amount of STDP activity per agent movement.  Parameters are chosen to strike a balance between system performance and evaluation time.  A typical SNN is shown in Fig.~\ref{khep}(a).

The SNNs were used to control a simulated Khepera II robot with 8 light sensors and 8 distance sensors.  At each timestep (64ms in simulation time), the agent sampled its light sensors, whose values ranged from 8 (fully illuminated) to 500 (no light) and IR distance sensors, whose response values ranged from 0 (no object detected) to 1023 (object very close).  All sensor readings were scaled to the range [0,1] (0 being unactivated, 1 being highly activated) before being input to the SNN.  Six sensors comprised the input state for the SNN, three IR and three light sensors at positions 0, 2, and 5 as shown in Fig.~\ref{khep}(b).  Additionally, two bump sensors were added to the front-left and front-right of the agent to prevent it from becoming stuck against an object.  If either bump sensor was activated, an interrupt was sent causing the agent to reverse 10cm and the agent to be penalised by 10 timesteps.  Movement values and sensory update delays were constrained by accurate modelling of physical Khepera agent.  Sensory noise was added based on Webots Khepera data; $\pm$2\% noise for IR sensors and $\pm$10\% noise for light sensors, all randomly sampled from a uniform distribution.  Wheel slippage was also included (10\% chance).  The spike trains of the output neurons were discretised into {\em high} or {\em low} activated ({\em high} activation if more than half of the 21 SNN processing steps generated a spike at the neuron,  {\em low} otherwise).  Three actions were possible: forward, (maximum movement on both left and right wheels, {\em high} activation of both output neurons) and continuous turns to both the left ({\em high} activation on the first output neuron, {\em low} on the second) and right ({\em low} activation on the first output neuron, {\em high} on the second) --- caused by halving the left/right motor outputs respectively.  Three discrete actions are used to encourage more distinct changes in network activity when generating and transitioning between these actions, allowing a detailed analysis of such disparities to be performed.

\subsubsection{Benchmark Synapse Types}
From the description of SNNs in Section~\ref{snn+esnn} and that of RMs in Section~\ref{RSM},  the strength of a connection weight in a neural network can be intuitively seen as the inverse resistance of that connection. The impact of a synapse on the functionality of the network depends on the modelling equations and parameters used.  In this section, the equations governing the two comparitive memristors (HP and PEO-PANI) and the constant connection are described.  The HP memristor was chosen for study as it is well understood.  The PEO-PANI memristor is also well-understood, but more importantly has a strongly different STDP profile (see Fig.~\ref{rsm-compare}(d)), which allows for contrasting behaviour.  Importantly when considering future hardware endeavours, both of these memristors are more likely than comparable devices to be available in sufficient quantities.

These synapse types have been previously compared~\citep{howardTEC}; the main findings of the study were (i) memristive STDP was used to generate highly fit solutions, (ii) the evolutionary algorithm assigned roles to the synapses based on their STDP behaviours (HP memristors were statistically more frequently connected to an inhibitory neuron, PEO-PANI were more frequently attached to an excitatory neuron), and (iii) self-adaptive search parameters were found to be context-sensitive and beneficial to the evolution of the networks.  In this study, these synapses serve as a means of comparison to the new variable devices.

{\bf HP Memristor}
\label{hp-mem}
The HP memristor is comprised of thin-film Titanium Dioxide (TiO$_{2}$) and Titanium Dioxide with oxygen vacancies (TiO$_{\mathrm{(2-x)}}$), which have different resistances.  The boundary between the two compounds moves in response to the charge on the memristor, which in turn alters the resistance of the device as delineated in Fig.~\ref{hpmem-vacancies}.  To allow for a future transition into variable synapses, memristance equations are rearranged based on the originals provided by~\cite{missing-mem-found}.  

In the following equations, $W$ is the scaled weight (conductance) of the connection, $G$ is the unscaled weight of the connection, $M$ is the memristance, $sf_1$ and $sf_2$ are scale factors, $R_{off}$ is the resistance of the TiO$_{2}$,  $R_{on}$ is the resistance of the oxygen-depleted TiO$_{(2-x)}$, $q$ is the charge on the device and  $q_{min}$ is the minimum allowed charge.  $\beta$ encompasses the physical properties of the device.  The original profiles, used for the benchmark memristors, are recreated using a rescaled $\beta=1$, $R_{on}=0.01$, $R_{off}=1$, $q_{min}=0.0098$.\\

$sf_1 = 0.99 /  \left( 1 - \left( \frac{\displaystyle1}{\displaystyle -R_{\mathrm{off}}  R_{\mathrm{on}}   \beta   q_{\mathrm{min}} + R_{\mathrm{off}}}\right)\right)$\hfill(5)\\

$sf_2 = 1 / \left(\frac{\displaystyle-R_{\mathrm{off}}    R_{\mathrm{on}}   \beta   ( R_{\mathrm{on}} - R_{\mathrm{off}})} {\displaystyle -R_{\mathrm{on}}   R_{\mathrm{off}}   \beta + R_{\mathrm{off}}}  sf_1\right) -1$\hfill(6)\\

$q =   \left(\frac{\displaystyle 1} {\displaystyle-R_{\mathrm{off}}  R_{\mathrm{on}}   \beta}\right)  \left( \frac{\displaystyle {\mathrm 1}}{ sf_1 (W + {\mathrm sf_2}) } - R_{\mathrm{off}}\right)$\hfill(7)\\

$M = -R_{\mathrm{off}}   R_{\mathrm{on}}   \beta   q + R_{\mathrm{off}}$\hfill(8)\\

$G = \frac{1}{M}$\hfill(9)\\

$W = G  sf_1 - sf_2 $\hfill(10)\\

{\bf PEO-PANI Memristor}
\label{peo-mem}
The polyethyleneoxide-polyaniline (PEO-PANI) memristor consists of layers of PANI, onto which Li$^+$-doped PEO is added ~\citep{peo-0}.  We have phenomenologically recreated the performance characteristics of the PEO-PANI memristor in terms of the HP memristor, creating a memristance curve similar to that reported by~\cite{peo-0}.  Two additional parameters, $G_{q_{\mathrm{min}}}$ and $G_{q_{\mathrm{max}}}$, are the values of $G$ when $q$ is at its minimum ($q_{\mathrm{min}}$) and maximum ($q_{\mathrm{max}}$) values respectively.  As with the HP equations, $\beta=1$ will produce the original PEO-PANI profile.\\

$q_{\mathrm{max}} = (R_{\mathrm{on}} - R_{\mathrm{off}}) / -R_{\mathrm{on}} R_{\mathrm{off}} \beta$\hfill(11)\\

$G_{q_{\mathrm{min}}} = 1/(-R_{\mathrm{off}}    R_{\mathrm{on}}   \beta    q_{\mathrm{min}} - R_{\mathrm{on}})+ R_{\mathrm{on}}$\hfill(12)\\

$G_{q_{\mathrm{max}}} = 1/(-R_{\mathrm{off}}    R_{\mathrm{on}}   \beta    q_{\mathrm{max}} - R_{\mathrm{on}})+ R_{\mathrm{on}}$\hfill(13)\\

The two scale factors are recalculated in (14) and (15).  Following this $q$ (16) and $M$ (17) are calculated, then G is calculated as in (9), and $W$ as in (10).\\

$sf_1 = 0.99 /( G_{q_{\mathrm{max}}} - G_{q_{\mathrm{min}}})$\hfill(14)\\

$sf_2 = (G_{q_{\mathrm{min}}}  sf_1) -0.01$\hfill(15)\\

$q =   \left(\frac{\displaystyle1}{\displaystyle -R_{\mathrm{off}} R_{\mathrm{on}} \beta}\right)\left(\frac{\displaystyle 1}{\displaystyle((W+sf_2)/sf_1)-R_{\mathrm{on}}} + R_{\mathrm{on}}\right) $\hfill(16)\\

$M = (-R_{\mathrm{off}}  R_{\mathrm{on}} \beta q - R_{\mathrm{on}}) + (1 / R_{\mathrm{on}})$\hfill(17)\\

{\bf Constant Connection}
\label{const-conn}
The constant connection possesses a static horizontal resistance profile, similar to a resistor.  The conductance of the connection is set randomly uniformly in the range [0,1] during initialisation and may be altered during application of the GA, but is unaffected by STDP and therefore constant during a trial.  We use the constant connection as a non-switching RSM, whose evolution is tailored via selection of a singular appropriate weight, rather than switching between two fixed weights. 

\subsubsection{Variable Synapse Types}
\label{variable-rsms}
Altering $\beta$ --- the memristor ``physical properties'' parameter --- allows the STDP responses of the MEM synapses to change.  The variable memristor profiles are allowed to switch between HP-like to PEO-PANI-like profiles, each of which are governed by their respective equations ((5)-(10) for HP, (11)-(17) for PEO-PANI).  Because of this, each memristor is augmented with a {\em type}, which is set to either 0 or 1 on memristor initialisation, with P=0.5 of each {\em type} being selected based on a uniform distribution.  If  {\em type = 0}, the refactored HP equations are used to calculate the profile of the device; otherwise the PEO-PANI equations are used.  $\beta$ is then initialised randomly uniformly in the range [$\beta_{{\mathrm min}}$, $\beta_{{\mathrm max}}$], where $\beta_{{\mathrm min}}=1$, $\beta_{{\mathrm max_{HP}}}=101$, and $\beta_{{\mathrm max_{PEO-PANI}}}=100$.  The combination of $type$ and $\beta$ is used to recreate the resistance profile of the device.

\begin{figure}[t!]
\begin{center}
\subfloat[]{ \epsfig{file=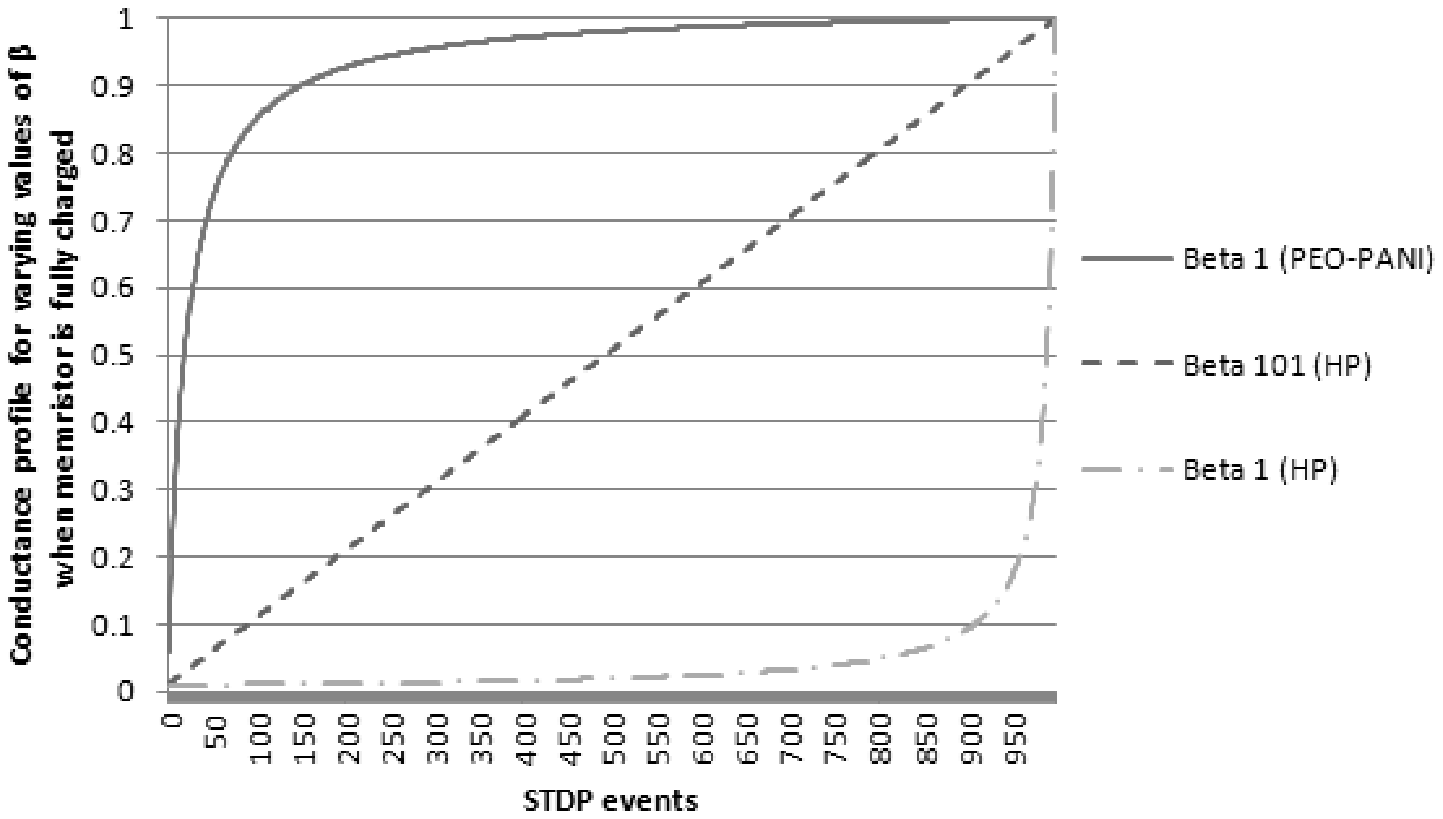,  height=2.5in, width=7cm}}
\subfloat[]{ \epsfig{file=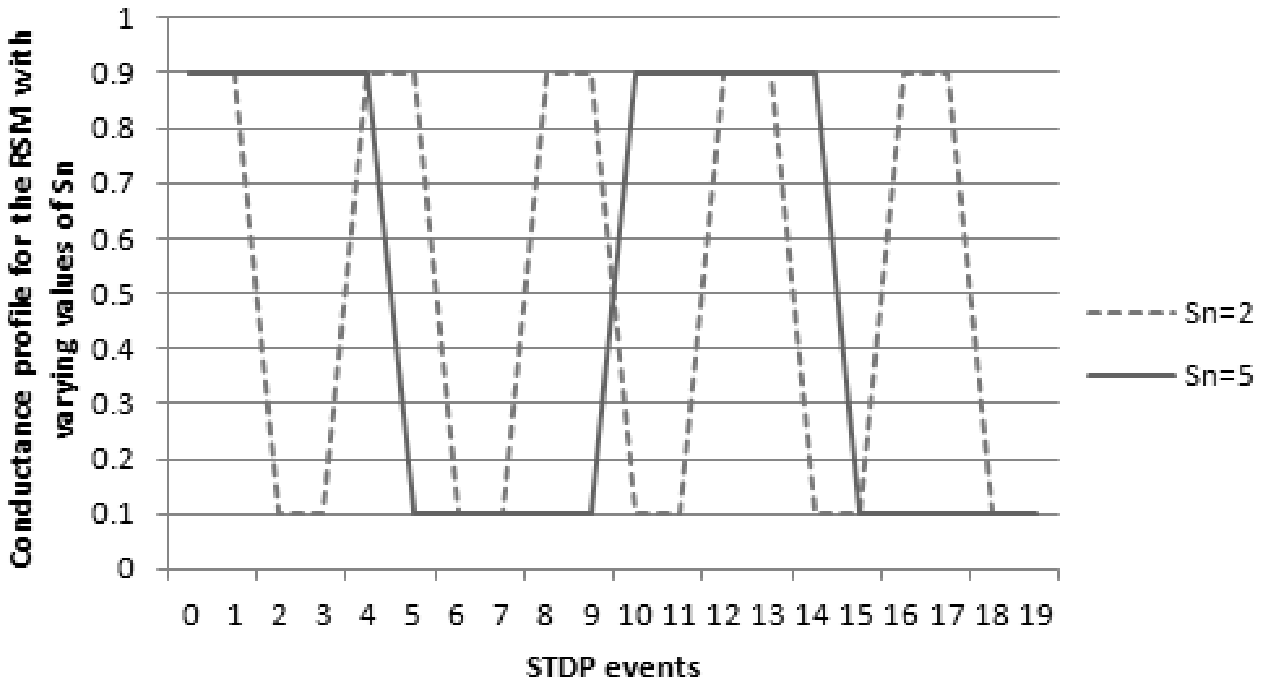, height=2.5in, width=6cm}}
\end{center}
\caption{(a) Displaying resistance profiles attained with different values of $\beta$ when fully charging a memristor. Here the $x$ axis shows 1000 positive STDP events assuming $L$=1000.  Static HP and PEO-PANI memristors have $\beta$=1. (b) Showing the resistance profile for an RSM with varying  $\theta_{LS}$ when supplied with constant STDP for 20 time steps.}
\label{profiles}
\end{figure}

STDP behaviour of the RSMs is described by $S_n$, which represents the number of consecutive STDP events required to cause a switch in an RSM (see Fig.~\ref{stdp-implementation}(b)).  A lower $S_n$ represents a higher sensitivity to voltage spikes and provides a higher maximum switching frequency --- see Fig~\ref{profiles}(b).  On synapse initialisation, the integer $S_n$ is selected uniform-randomly in the range [$S_{{\mathrm n_{min}}}$,$S_{n_{{\mathrm max}}}$], so that a minimum of $S_{n_{{\mathrm min}}}$ consecutive STDP events are required to cause a switch.  RSM parameters were $S_{n_{{\mathrm min}}}$=2, $S_{n_{{\mathrm max}}}$=6.  

\subsubsection{STDP Implementation}
\label{stdp}

Section~\ref{plasticity} reviewed a number of STDP implementations.  We follow~\cite{stdp-discrete} and~\cite{nano-mem-syn-neuromorphic} in using discrete-time stepwise waveforms, as our SNNs operate in discrete time.  Each neuron in the network is augmented with a ``last spike time'' variable $LS$, which is initially 0.  When a neuron spikes, this value is set to an experimentally-determined positive number.  At the end of each of the 21 steps that make up a single timestep, each RSM is analysed by summating the $LS$ values of its presynaptic and postsynaptic neurons.  Following this, each $LS$ value is then decreased by 1 to a minimum of 0, creating a discrete stepwise waveform through time.

\begin{figure}[t!]
\begin{center}
\subfloat[]{\epsfig{file=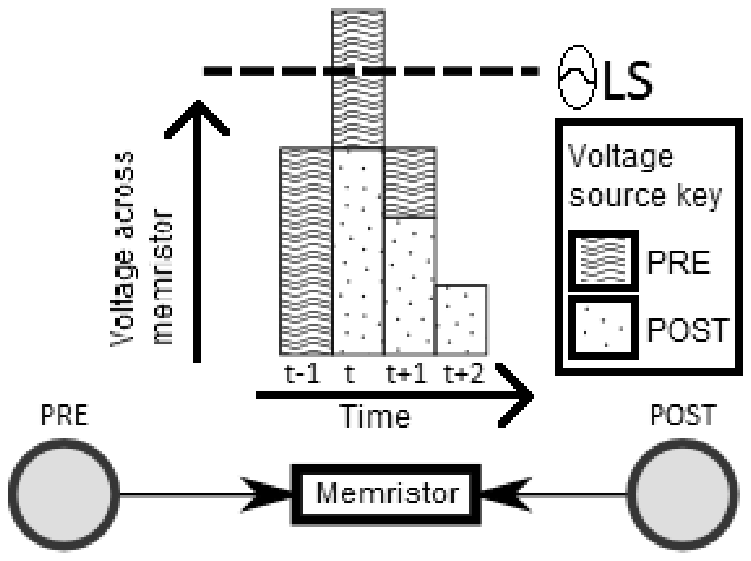,  height=5cm, width=5cm}}
\subfloat[]{ \epsfig{file=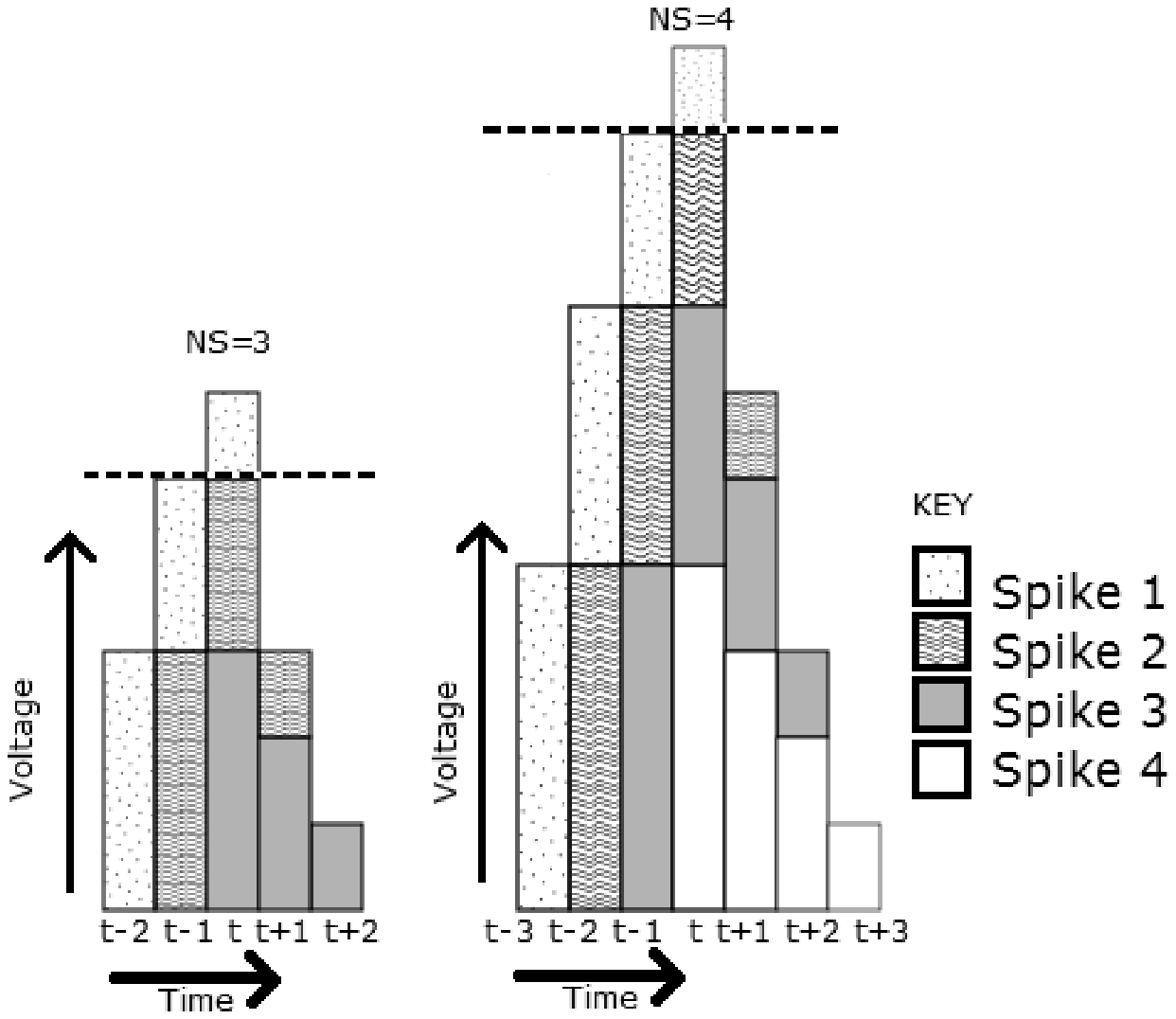, height=6cm, width=7cm}}
\end{center}
\caption{Showing a positive STDP event for a memristor.  A presynaptic voltage spike is received at time {\em t-1}, with a postsynaptic voltage spike at time {\em t}.  Combined, the voltage surpasses $\theta_{LS}$, increasing the conductivity of the device.  (b) In the case of the RSM, consecutive voltage spikes (l.h.s $S_n=3$, r.h.s $S_n=4$) serve to push the voltage past a threshold, causing a switch.  Dotted lines show the derived voltage threshold.  Voltage spike values are decremented by one per subsequent time step.}
\label{stdp-implementation}
\end{figure}

For memristors, all connection weights are initially 0.5 and can vary in-trial via STDP.   If $LS$ exceeds the positive threshold $\theta_{LS}$, the efficacy of the synapse changes (see Fig.~\ref{stdp-implementation}(a)).  The synapse increases or decreases its efficacy depending on which neuron has the highest $LS$ value, providing pre- to postsynaptic temporal coincidence.  If the $LS$ values are identical, STDP does not occur.  Each STDP event either increases or decreases $q$ by $\Delta q$, as detailed in (18), which is then used to calculate memristor conductivity as detailed in (5)-(10) for HP memristors, or (11)-(17) for PEO-PANI memristors.  Parameter $L$, the number of steps to take the memristor from fully resistive to fully conductive, is set to 1000.  Additionally, $LS=3$, $\theta_{LS}=4$\\

$\Delta q = (q_{\mathrm{max}}-q_{\mathrm{min}}) / L $\hfill(18)\\
\\\\\\
Fig.~\ref{profiles}(a) shows that HP and PEO-PANI memristors display increased sensitivity (larger $\Delta W$ per STDP event) when $\beta$ is a low number and either $W>0.1$ for HP-governed synapses, or $W<0.9$ for PEO-PANI-governed profiles.  

RSM STDP parameters are $S_n$, which encapulates the sensitivity of the device to voltage buildup (in the form of repeated STDP spikes), and $S_c$, which tracks the number of consecutive STDP events the synapse has experienced.  All synapses are initially in the Low Resistance State (W=0.9).  At each step every synapse is checked as before, incrementing $S_c$ if an STDP event occurs at the synapse and decrementing $S_c$ if no STDP event occurs at that step.  If $S_n$=$S_c$, the RSM switches to the High Resistance State ($W=0.1$) and $S_c$ is reset to 0.  The RSM can switch back and forth between the LRS and HRS during a trial.  Note that the polarity of the voltage spike in RSM networks is always the same, regardless of the coincidence of presynaptic and postsynaptic spikes across the device.    

Fig.~\ref{stdp-implementation}(b) shows how this mechanism compares to regular STDP.  Due to the requirement of consecutive STDP events, the actual frequency of synapse alteration is lower than that seen in the memristive networks.  The number of consecutive spikes required can be seen as an analogue to a higher voltage threshold which is required to switch.  Note that $q$ is no longer instrumental to the functioning of the device; i.e. RSMs are related to voltage-threshold rather than being charge-controlled.  Because of this, the network architectures required to handle the two Resistive Memories (memristive and RSM) are assumed to be incompatible with each other.

We note that $LS$ may be discussed in the context of RSMs and must be set equal to $S_n$ to allow the device to surpass this threshold in the face of voltage dissapation, which acts to lower the value of the waveform through time.     $\theta_{LS}$ may also be derived and varies according to $S_n$, although its value is not necessarily required to model the device.  After an RM network is trialled, all synapses are reset to their original weights (0.5 for memristors, 0.9 for RSMs).

\subsection{Evolutionary Algorithm}
\label{discovery}
Having described the component parts of our networks, we now detail the implementation of the GA that acts on them.  Per generation, two parents are selected fitness-proportionately, mutated, and used to create two offspring.  We use only mutation to explore weight space; crossover is omitted as sufficient solution space exploration can be obtained via a combination of self-adaptive weight and topology mutations; a view that is reinforced in the literature (e.g.,~\cite{mu-only-needed}).  The offspring are inserted into the population and two networks with the lowest fitness deleted.

Each network is respresented two variable-length vectors, one containing neurons and the other connections.  A neuron is defined by its type (excitatory or inhibitory), membrane potential, and ``last spike'' value $LS$.  A connection is defined by its type (e.g. HP memistor),  weight, charge, $\beta$ / $S_{n}$, and the neurons it connects.  These two vectors are augmented by self-adaptive parameters that control various rates of mutation.  Mutatable network parameters are neuron type, synaptic weight (in non-RM networks), $\beta$ (in variable memristor networks), $S_{n}$ (in variable RSM networks), and associated self-adaptive parameters.  Neurons and connections may be added/removed from their respective vectors by the GA.

The use of a self-adaptive framework is justified when the application area of neuromorphic computing is considered --- brainlike systems must be able to autonomously adapt to a changing environment and adjust their learning accordingly.  This potentially allows increased structural stability in highly fit networks whilst enabling less useful networks to vary more strongly per GA application.  Mechanistically, self-adaptation also permits the use of self-repair/self-modification, wherein the GA  is able to (i) in a stable environment, lower mutation rates to enable homeostatis or provide incremental, gradual improvements or (ii) when the environment rapidly changes, or part of the network fails, increase learning rates to more quickly adapt to these new conditions.  When coupled with neuroevolution, the effect is to tailor the evolution of the network to the complexity of the environment explicitly, e.g. each network controls its own architecture autonomously in terms of (i) amount of mutation that takes place in a given network at a given time (ii) adapting the hidden-layer topology of the neural networks to reflect the complexity of the problem considered by the network, as shown by~\cite{HandB}.  This mechanism was first used with SNNs by~\cite{howard}.  We note that benefits of this approach will be more pronounced in hardware implementations, which will be the topic of future research.

\subsubsection{Self-adaptive Mutation}
\label{sa-mutation}
We use self-adaptive mutation rates as in Evolutionary Strategies (ES)~\citep{rechenberg}, to dynamically control the frequency and magnitude of mutation events taking place in each network.  Here, $\mu$ $(0<\mu\leq1)$ (rate of mutation per allele) of each network is initialized uniformly randomly in the range [0,0.25]. During a GA cycle, a parent's $\mu$ is modified as in (19) (where $N$ denotes a normal distribution); the offspring then adopts this new $\mu$ and mutates itself by this value, before being inserted into the population.\\

$\mu \rightarrow \mu \exp^{N(0,1)}$ \hfill(19)\\

Only non-RM networks can alter their connection weights via the GA.  Connection weights in this case are initially set during network creation, node addition, and connection addition randomly uniformly in the range [0,1].  Memristive network connections are always set to 0.5, and cannot be mutated from this value.  This aims to force the memristive networks to harness the plasticity of their connections during a trial to successfully solve the problem.

\subsubsection{Control of Variable Synapses}
The STDP responses of the variable synapses are governed by the self-adaptive parameter $\iota$, which is initialised and self-adapted as with $\mu$.  In the case of the variable memristor, mutation changes a synapse's $\beta$ by $+/-$10\% of the total range of $\beta$.  If a memristor's new value of $\beta$ surpasses a threshold $\beta_{{\mathrm max}}$, the {\em type} of the memristor is switched and a new $\beta$ calculated as $\Delta\beta$ - $\beta_{{\mathrm max}}$.  In this way, a smooth transition between the different profile types is provided.  For the variable RSM, mutation alters $S_n$ by $\pm1$ of its current value, constrained to the range [$S_{n_{{\mathrm min}}}$,$S_{n_{{\mathrm max}}}$].  Whereas memristive STDP can be viewed as a form of in-trial context-sensitive weight mutation, as shown in Fig.~\ref{profiles}(a), RSM STDP is more akin to context-sensitive connection selection (Fig.~\ref{profiles}(b)). 

\subsubsection{Topology Mechanisms}
\label{topology}
Given the desire for adaptive solutions, it would be useful if appropriate network structure is allowed to develop until some task-dependent required level of computing power is attained.  A number of encoding variants have been developed specifically for neuroevolution, including Analog Genetic Encoding (AGE)~\citep{4336123}, which allows for both neurons and connections to be modified,  amongst others~\citep{4016064}.  A popular framework is NeuroEvolution of Augmenting Topologies (NEAT)~\citep{neat} which combines neurons from a predetermined number of niches to encourage diverse neural utility and enforce niche-based evolutionary pressure.  This method has been shown to be amenable to real-time evolution~\citep{1545941}.  

Two topology morphology schemes allow the modification of the spiking networks by (i) adding/removing hidden-layer nodes, and (ii) adding/removing inter-neural connections.  Each network has a varying number of hidden-layer neurons (initially 9, and always $>$ 0).  Additional neurons can be added or removed from the single hidden layer based on  two self-adaptive parameters $\psi$ $(0<\psi\leq1)$ and $\omega$ $(0<\omega\leq1)$. Here, $\psi$ is the probability of performing neuron addition/removal and $\omega$ is the probability of adding a neuron; removal occurs with probability $1- \omega$. Both have initial values taken from a random uniform distribution, with ranges [0,0.5] for $\psi$ and [0,1] for $\omega$.  Offspring networks have their parents $\psi$ and $\omega$ values modified using (19) as with $\mu$, with neuron addition/removal occurring after mutation.  Added nodes are initially excitatory with 50\% probability, otherwise they are inhibitory.

Automatic feature selection is a method of reducing the dimensionality of the data input to a process by using computational techniques to select and operate exclusively on a subset of inputs taken from the entire set.  In the context of neural networks, feature selection can disable synaptic (traditionally input) connections.  In this study we allow each connection to be individually enabled/disabled, a mechanism termed ``Connection Selection''.  During a GA cycle a connection can be added or removed from the connection vector based on a new self-adaptive parameter $\tau$ (which is initialized and self-adapted in the same manner as $\mu$ and $\psi$).  If a connection is added for a non-memristive network, its connection weight is randomly initialised uniformly in the range [0,1], memristive connections are always set to 0.5.  During a node addition event, new connections are set probabilistically, with $P = 0.5$ of each possible neural connection being added.  Connection selection is particularly important to the memristive networks. As they cannot alter connection weights via the GA, variance induced in network connectivity patterns plays a large role in the generation of STDP in the networks. Likewise, RSM networks rely on Connection Selection to generate synchronised synaptic excitations/inhibitions, which allow the network to generate appropriate output actions.

\begin{figure}[t!]
\begin{center}
\subfloat[]{ \epsfig{file=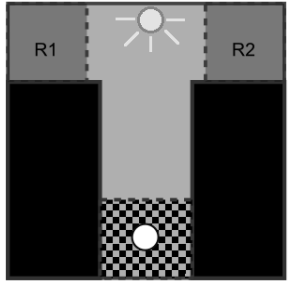,width=1.5in ,height=1.5in}}
\end{center}
\caption[]{The T-maze.  The agent (white circle) begins in the checkered box and must travel to R1 or R2.  The light source is shown (top-centre).}
\label{env}
\end{figure}

\subsection{Task}
The main advantage of STDP is the ability to vary during a trial in response to a dynamically-changing environment.  To demonstrate the ability of variable RM synapses to handle dynamic reward scenarios, an experiment was simulated in a T-maze (e.g.~\citep{tmaze}) scenario, as used by~\cite{solt-dyn} to investigate the adaptivity of plastic networks.  In particular, the variable memristor (MEM) and RSM (RSM) elements were compared to the static HP (HP) and PEO-PANI (PEO) memristors, and the constant connection or nonswitching RSM (CONST).  The popular robotics simulator Webots ~\citep{webots04} was chosen; alternatives are summarised by~\cite{robot-sim-survey}.

The environment was an enclosed arena with coordinates ranging from [-1,1] in both $x$ and $y$ directions, and is shown in Fig.~\ref{env} to represent a ``T''.  The agent is initialised facing North in a zone at the bottom of the ``T'' (delineated in Fig.~\ref{env} with a checkered pattern).  Reward zones R1 and R2, were situated at the end of the left and right arms respectively.  A light source, modelled on a 15 Watt bulb, was placed at the top-centre of the arena ($x=0.5$, $y=1$) and was used by the network for action calculation.  

During a trial, the agent initially learned to navigate to R1.  Once stable pathfinding to R1 was attained, the reward zone was switched to R2 and the agent reinitialised in the start zone.  To give the agent ``memory'' of the first part of the trial, the membrane potentials and synaptic weights of the controlling network were not reset during this process.  This task is dynamic as the reward zone changes so the agent must ``forget'' it's previously-learned behaviour after a time and adapt to a newly-positioned goal state.   A network that located R2 following location of R1 was said to have {\em solved} the trial,  The measure of fitness used,  $f$,  was simply the total number of steps required to solve the trial.  Each network was permitted 4000 steps to locate R1, plus an additional 4000 steps to locate R2.  

\begin{figure}[ht!]
\begin{center}

\subfloat[]{ \epsfig{file=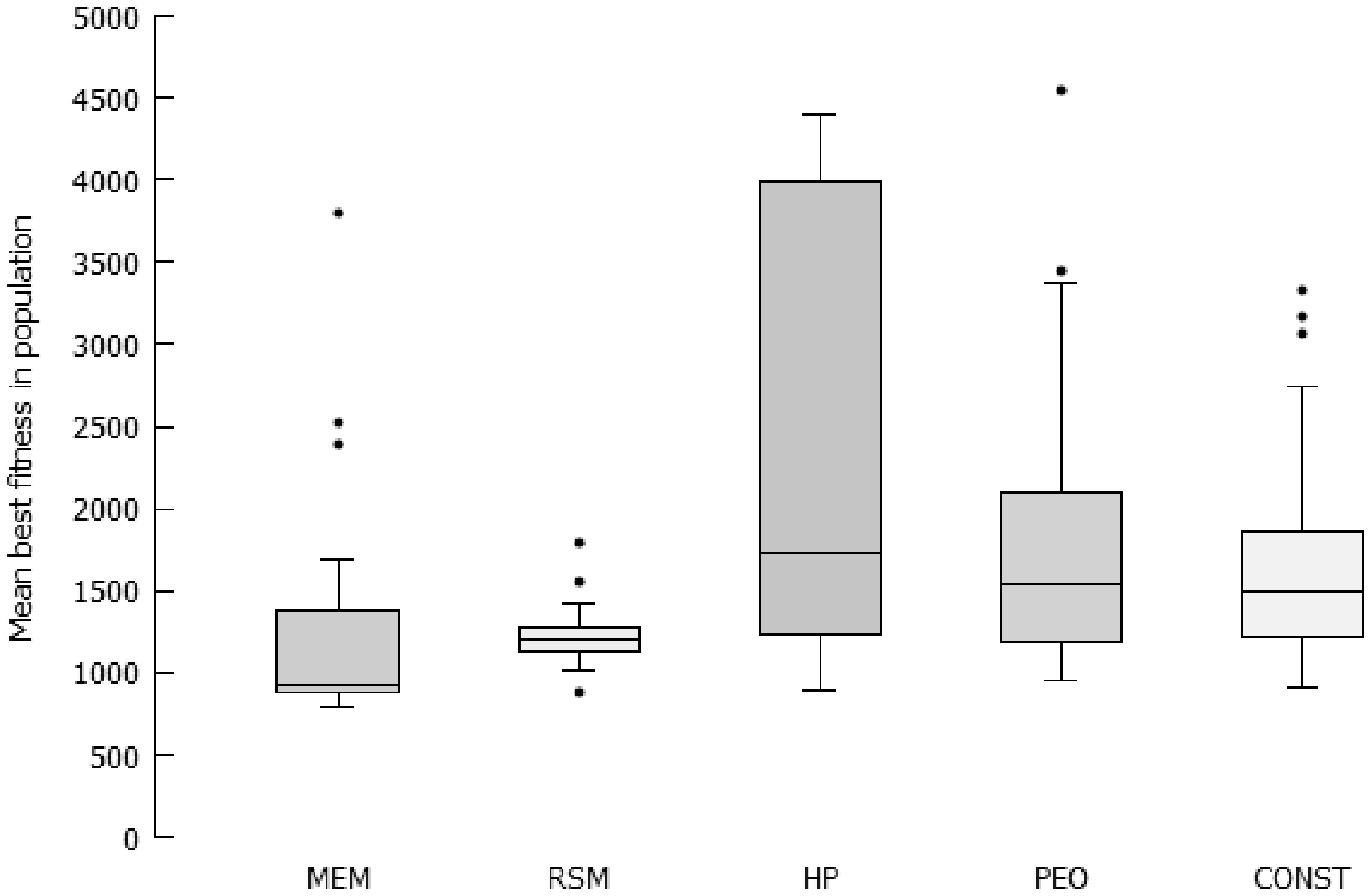,width=6.2cm ,height=3cm}}
\subfloat[]{ \epsfig{file=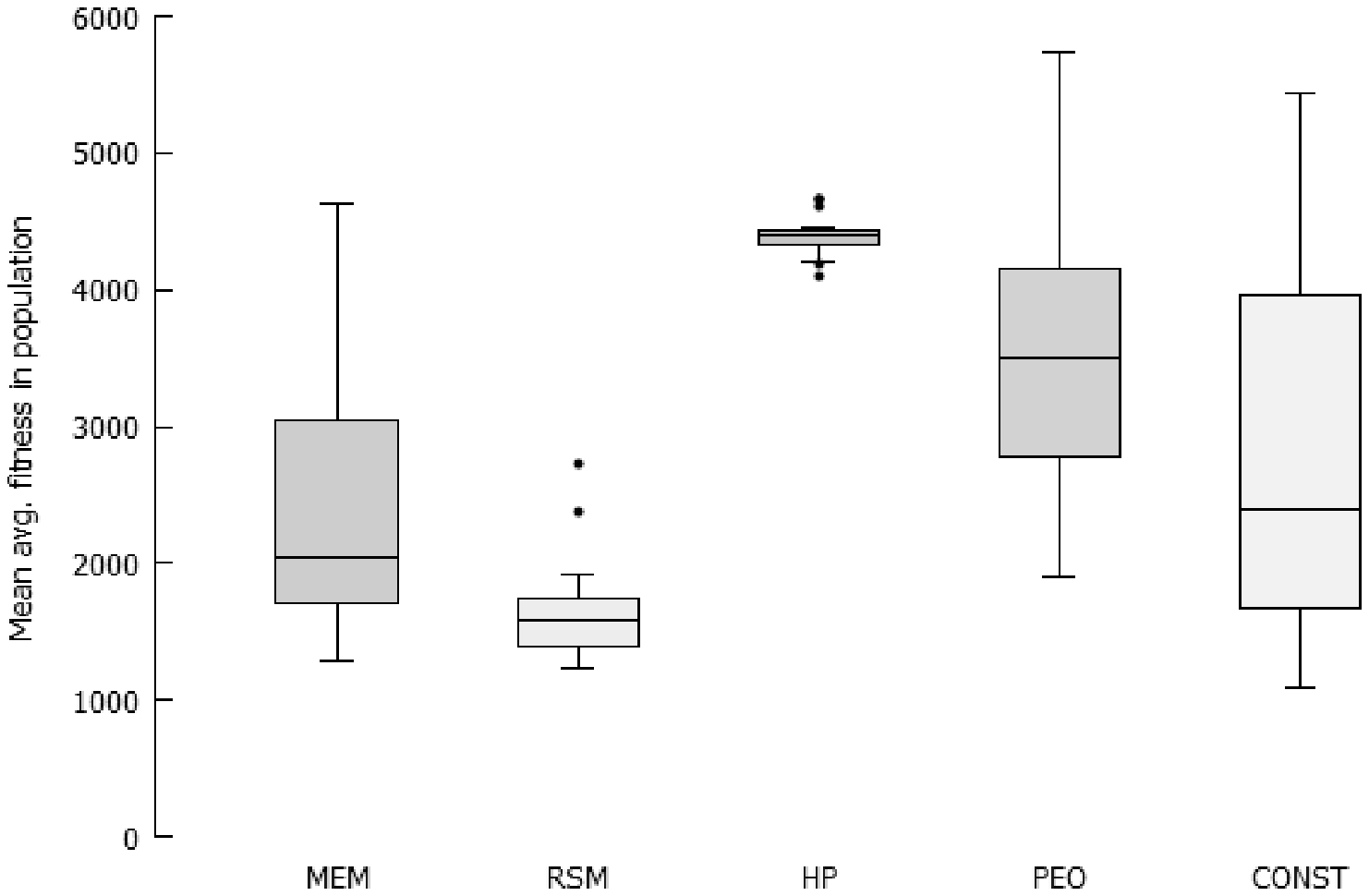,width=6.2cm,height=3cm}}\\
\subfloat[]{ \epsfig{file=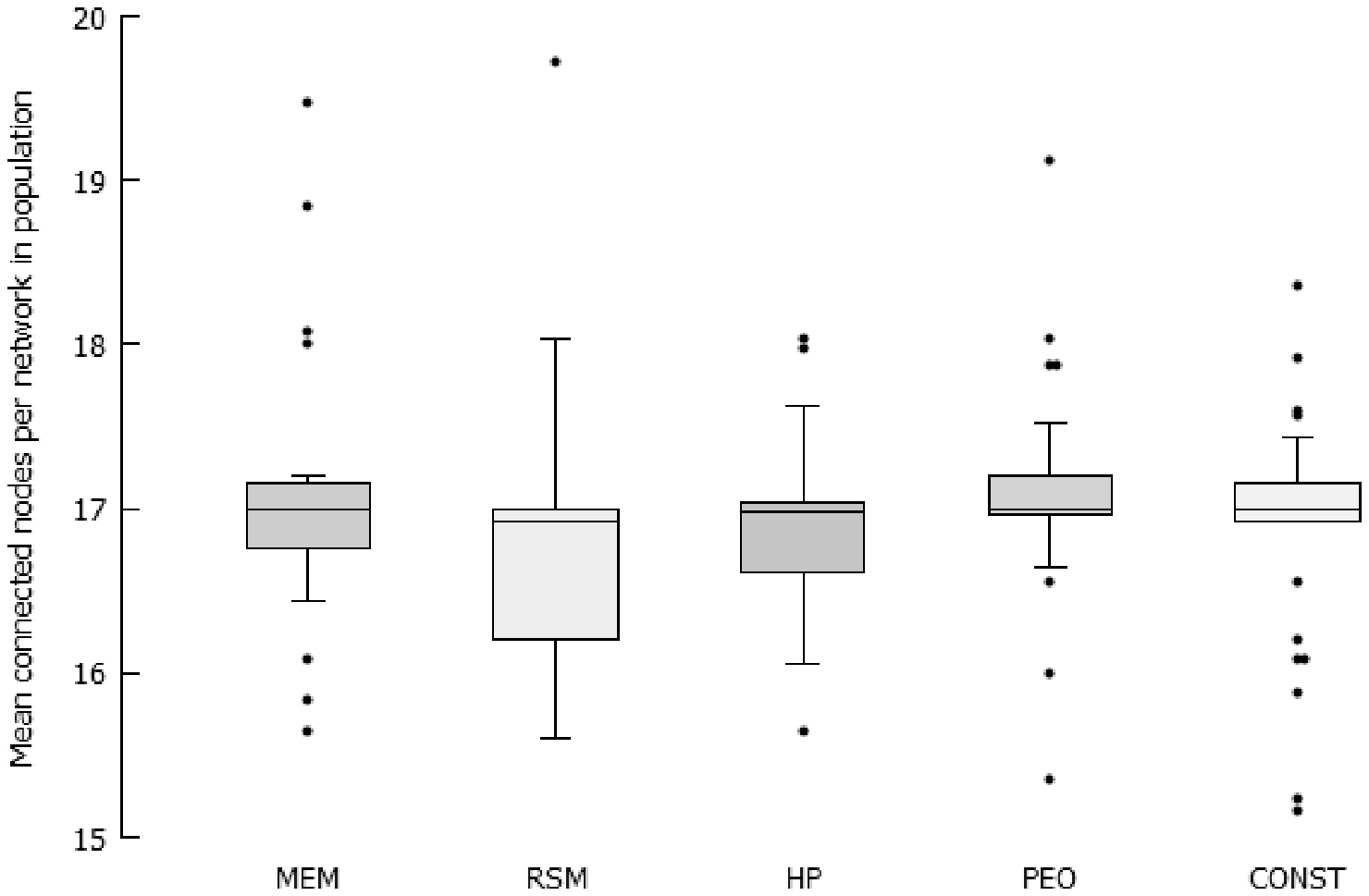,width=6.2cm,height=3cm}}
\subfloat[]{ \epsfig{file=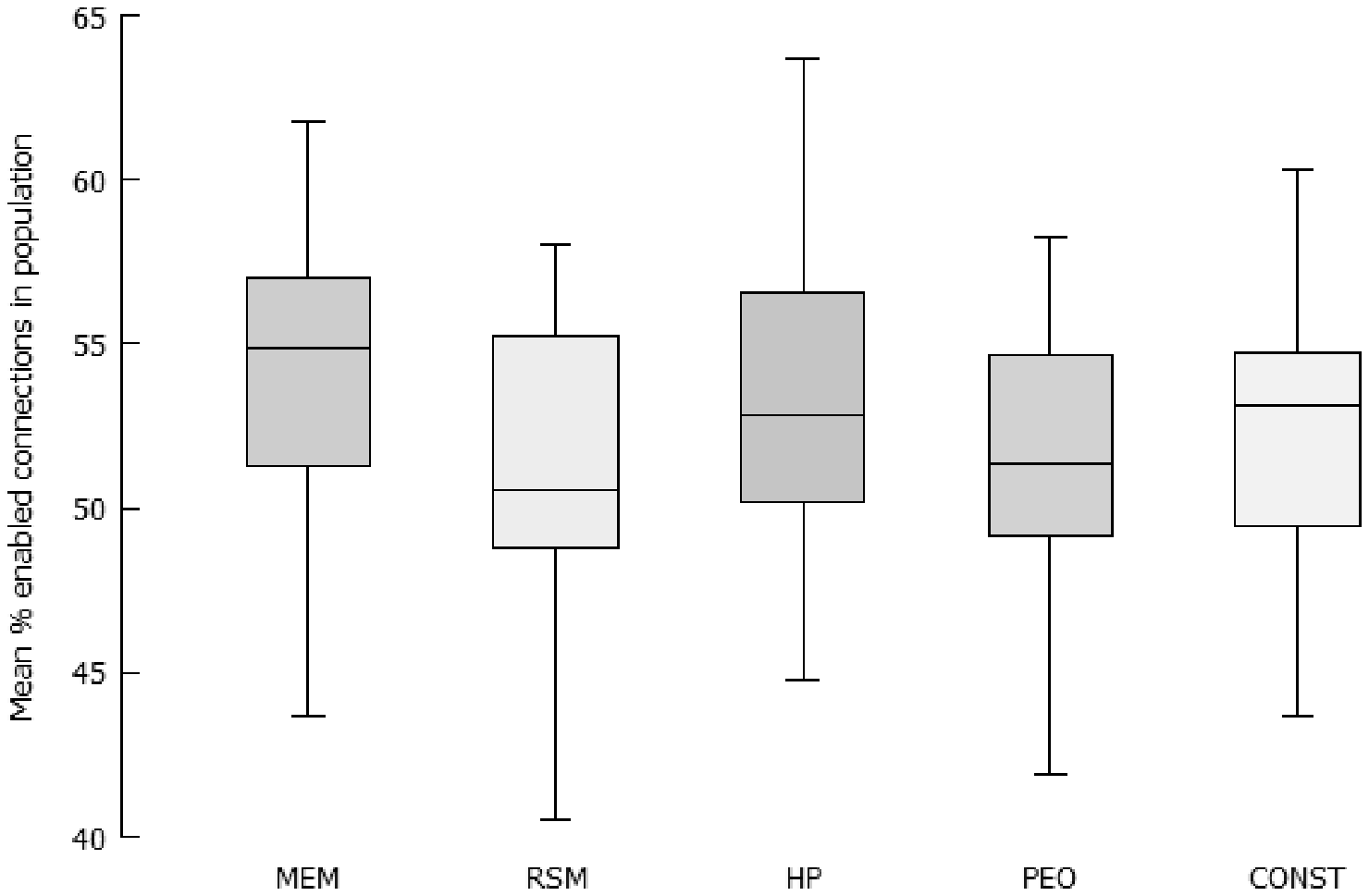,width=6.2cm,height=3cm}}\\
\subfloat[]{ \epsfig{file=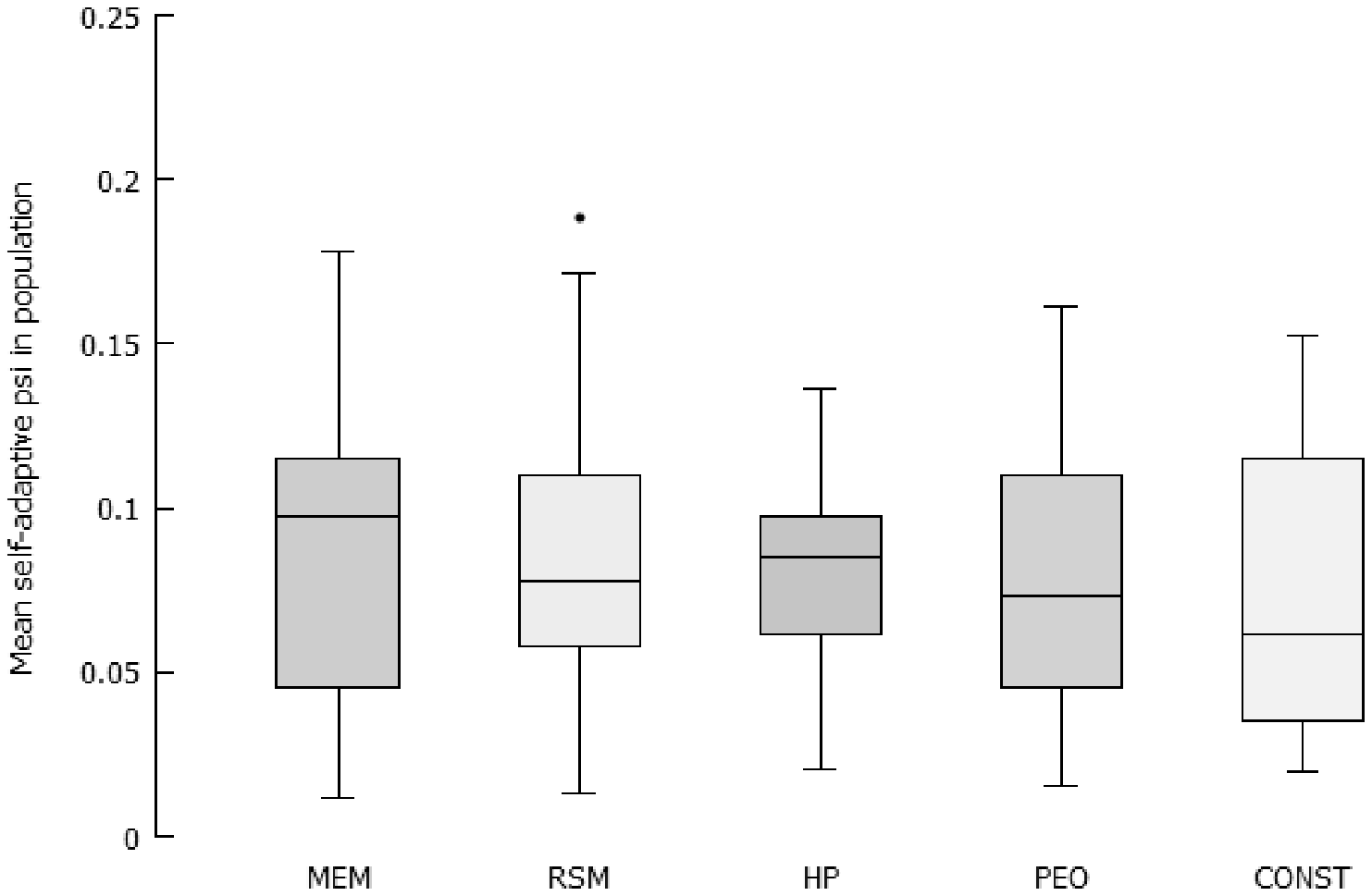,width=6.2cm,height=3cm}}
\subfloat[]{ \epsfig{file=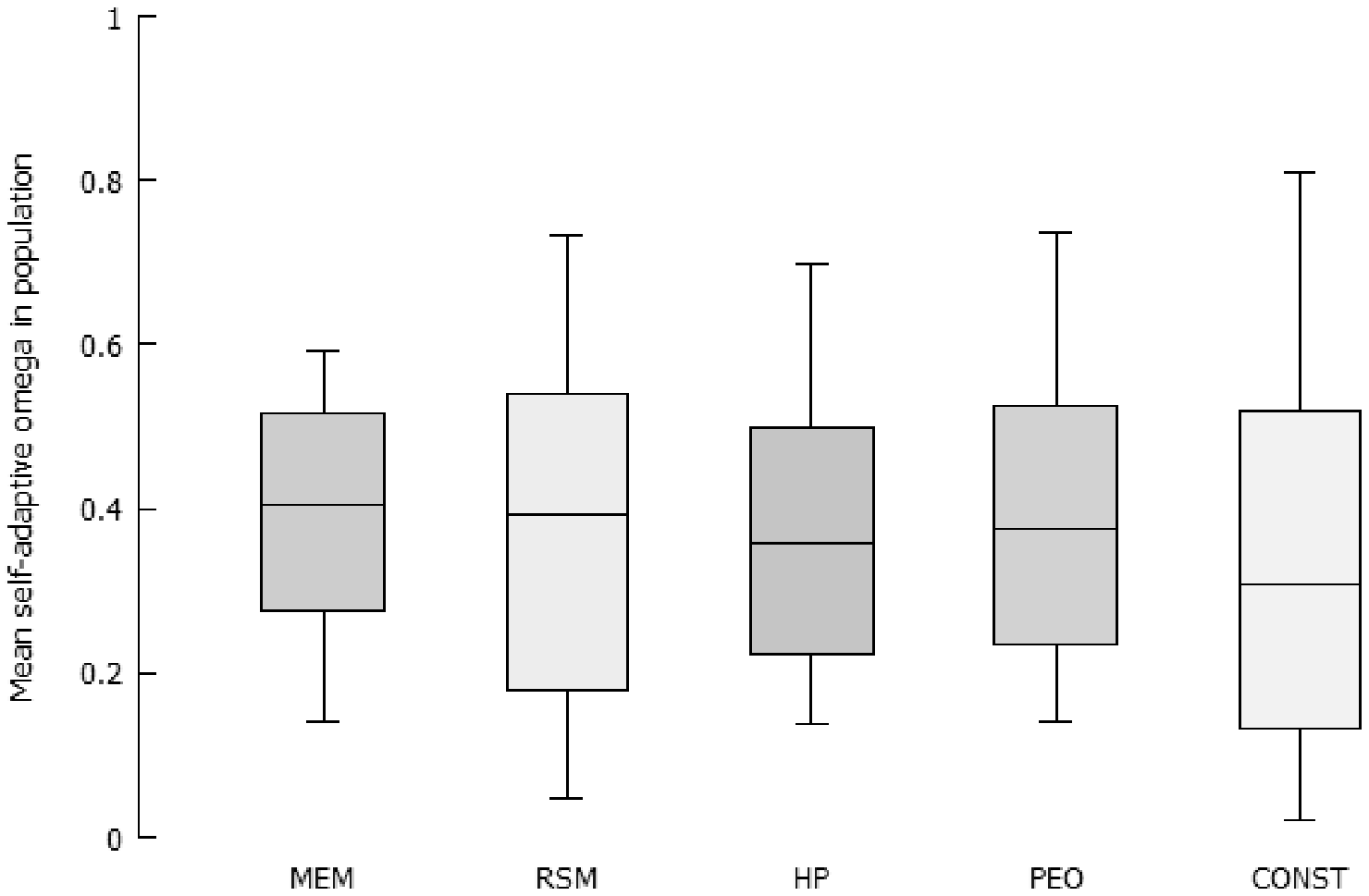,width=6.2cm,height=3cm}}\\
\subfloat[]{ \epsfig{file=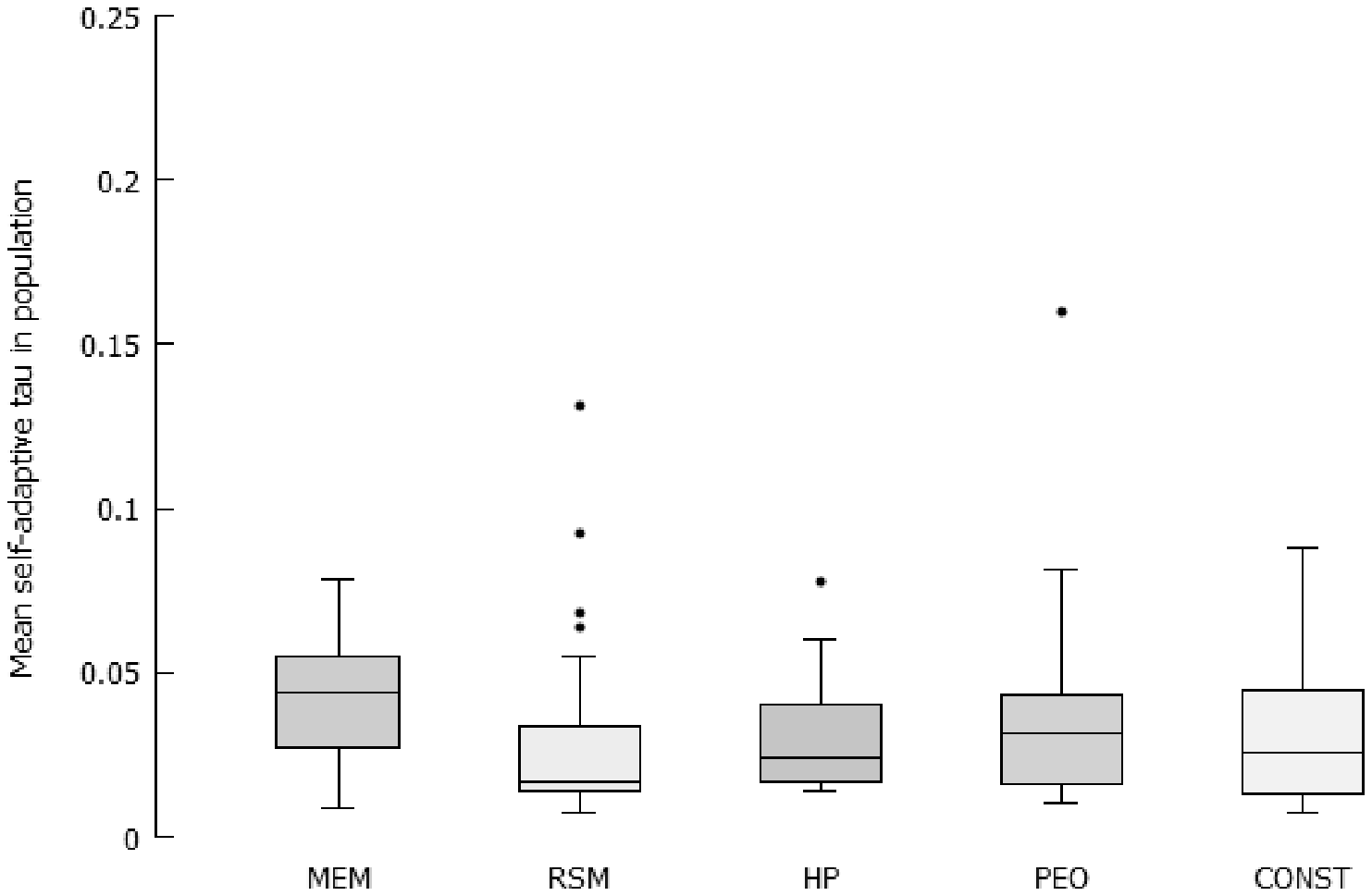,width=6.2cm,height=3cm}}
\subfloat[]{ \epsfig{file=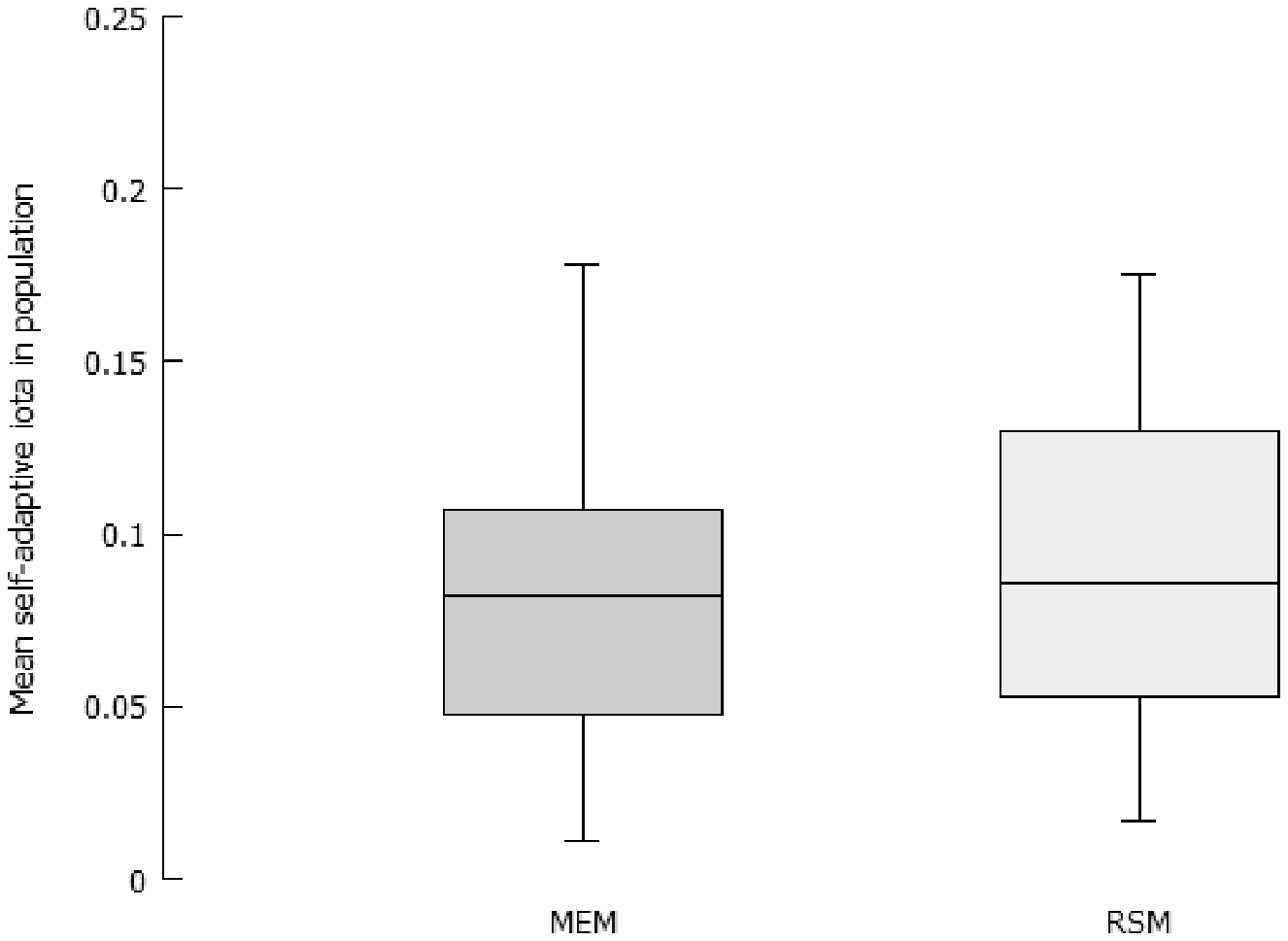,width=6.2cm,height=3cm}}\\

\end{center}
\caption[]{Box plots showing Average (a) best fitness (b) mean fitness (c) connected hidden -ayer nodes (d) enabled connections (e) $\psi$ (f) $\omega$ (g) $\tau$  (h) $\iota$ for the T-maze experiment.}
\label{dyn-graphs}
\end{figure}

\section{Results}

In the following discussion, ``Best fitness'' refers to the lowest-fitness network in each experimental repeat (lower fitness denoted higher solution quality).  ``Average fitness'' was the mean fitness of the population of networks.``Neurons'' was the average final number of connected neurons in the population and ``Connectivity'' was the average percentage of enabled connections in the population.  All of the above measurements were averaged over the 30 experimental repeats.  Two-tailed T-tests were used to asses statistical significance.

Figs.~\ref{dyn-graphs}(a) and (b) show that both variable network types generate solutions with better average fitness and best fitness than their static counterparts (for best fitness, MEM and RSM were both $p<$0.05 vs. HP, PEO, and CONST; for average fitness MEM was $p<$0.05 vs. HP and PEO, RSM was $p<$0.05 vs. HP, PEO, and CONST).  This is an encouraging result, as in addition to outperforming the majority of the non-variable synapses,  the variable Resistive Memory synapses induce no significant performance overhead that may have arisen due to the increased search/behaviour space that the GA has to deal with.  The average fitness of HP networks is due to much of the population not being able to perform both parts of the trial; best fitness is still comparable to the other non-variable network types.  When comparing the two variable RM networks, we note that the average fitness of RSM networks is significantly (p$<$0.05) lower than that of MEM networks. 

The best single network of each type had the following fitness values:  MEM=839, RSM=879, PEO=956, CONST=915, HP=901.  MEM networks generated the highest quality (single) solution, 2.56 seconds faster than the best-evolved RSM network.  As the pathfinding function is better-approximated, GA-tuned STDP behaviour of the MEM synapses is said to be capable of the most complex responses to environmental stimuli.

Variable RMs are also found to be easier to evolve for this task  - both MEM and RSM networks take statistically (p$<$0.05) fewer generations to generate a solution that successfully solves the task compared to all other synapse types, indicating that a greater rate of adaptation is possible when using variable synapses.  The average number of generations to create a network that solves the problem are RSM=19.3, MEM=27.7, PEO= 46.2, HP=890.5, and GA=66.3.  Compared to RSM networks, MEM networks appear to require more GA tuning of $\beta$ to generate functional solutions, due to the larger and more fine-grained search space of $\beta$ compared to $S_n$. 

RSM networks possess the fewest average neurons per network.  MEM networks contain more excitatory neurons (average 13.9) than they do inhibitory (average 3.2, p$<$0.05), the most likely explanation being that a given level of network activity is required to generate all required behaviours given an arbitrary input state.  RSM networks possess an approximately equal distribution of neuron types because the neurons are ambivalent to the polarity of the incoming voltage spikes (average 6.7 excitatory neurons, 5.9 inhibitory neurons). 

MEM networks were more densely connected (average 53.6\%) than their RSM counterparts (average 50.9\%, p$<$0.05).  Denser synaptic topologies were required for MEM networks as more of the computational power of the network is embodied in the synapse itself (a notion echoed in recent literature~\citep{synaptic-computation}).  When a single MEM synapse is compared to a single RSM synapse, it is evidenced that the MEM device is capable of temporal behaviour that is more complex and granular, however the RSM network as a whole overcomes a lack of synaptic complexity via arrangements of simpler devices and synchronised LRS/HRS switching to generate useful spike patterns.  HP networks are also more densely connected than RSM (average 53.3\%, p$<$0.05), although this is likely due to the requirement for additional synapses to percolate sufficient activity through the network given the tendancy of the HP memristor to either become stuck in a low-conductivity region of its STDP profile or be used as a depressing synapse e.g. with a presynaptic inhibitory neuron (\cite{howardTEC} provide in-depth analysis).

In terms of computational complexity, we observe that MEM and RSM networks possess more complex temporal dynamics than the other STDP-enabled network types.  Complexity in MEM networks is evidenced through a heterogeneous collaboration of myriad STDP responses.  In contrast, complexity in RSM networks is mainly due to the creation of synchronised weight oscillators within the network which are required for high-fitness behaviour generation. In either case, synaptic activity (as opposed to neuron activity) is seen to be the main driver of action generation --- the computational onus is seen to be taken from neuron and reattributed to the synapse.  

Connection Selection approximately halves the amount of connections used by the networks (Fig.~\ref{dyn-graphs}(d) shows connectivity varying between 50.9\% --- 53.7\%).  Considering possible NC applications, the integration of mechanisms such as synaptic redundancy, self-repair, and self-modification could be eased by the sparse connectivity of the generated solutions, e.g. in eventual hardware implementations, more synapses will be available to implement these mechanisms when compared to solutions that do not use this technique.

Figs.~\ref{dyn-graphs}(e) --- (h) shows that all parameters decline from their original values.  MEM networks are seen to possess higher average $\tau$ (rate of Connection Selection, 0.043) than both RSM (0.029), HP (0.03) and CONST (0.033) networks (p$<$0.05).  Whilst proving the context-sensitivity of self-adaptation, this result also relates to the idea that MEM synapses are more computationally powerful, therefore the networks (i) require more connections (ii) need to experiment more with synaptic configurations via Connection Selection to generate highly fit solutions.  Iota, which controls the frequency of variable synapse alteration events, varies between MEM and RSM with p=0.481.

When self-adaptive parameters from the same network type are compared, all are observed to be significantly (p$<$0.05) different from each other --- compare Fig.~\ref{dyn-graphs}(e)-(h).  Self-adaptation is shown to be context sensitive, with the GA automatically discovering preferential values for each parameter based on its role in the generation of fit solutions.

\begin{figure}[ht!]
\begin{center}

\subfloat[]{ \epsfig{file=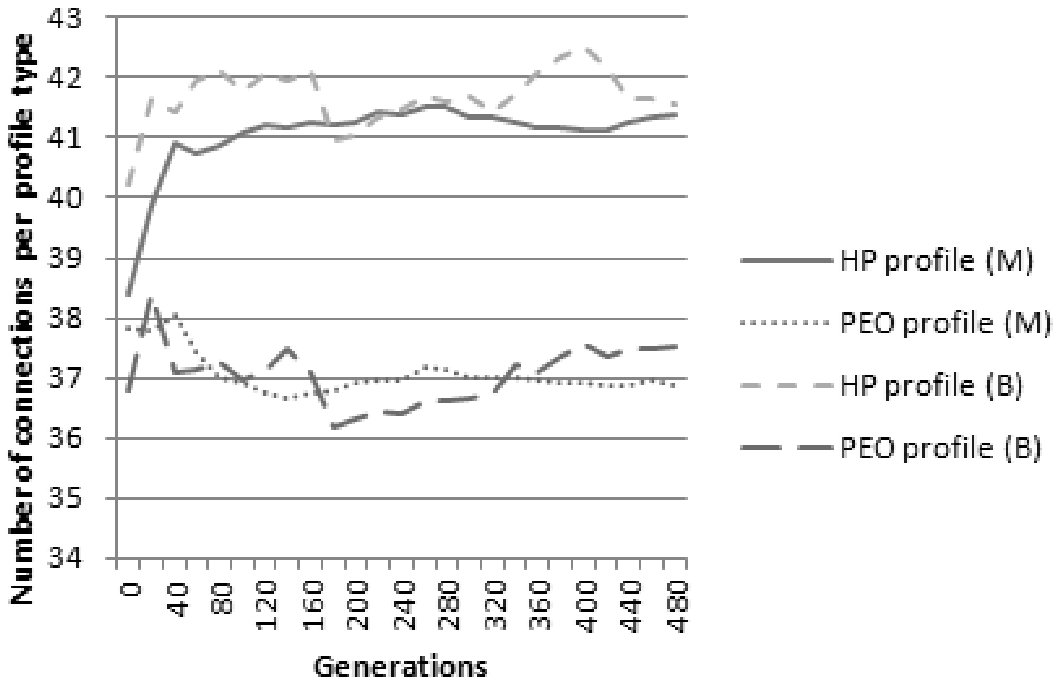,width=6.5cm,height=3.5cm}}
\subfloat[]{ \epsfig{file=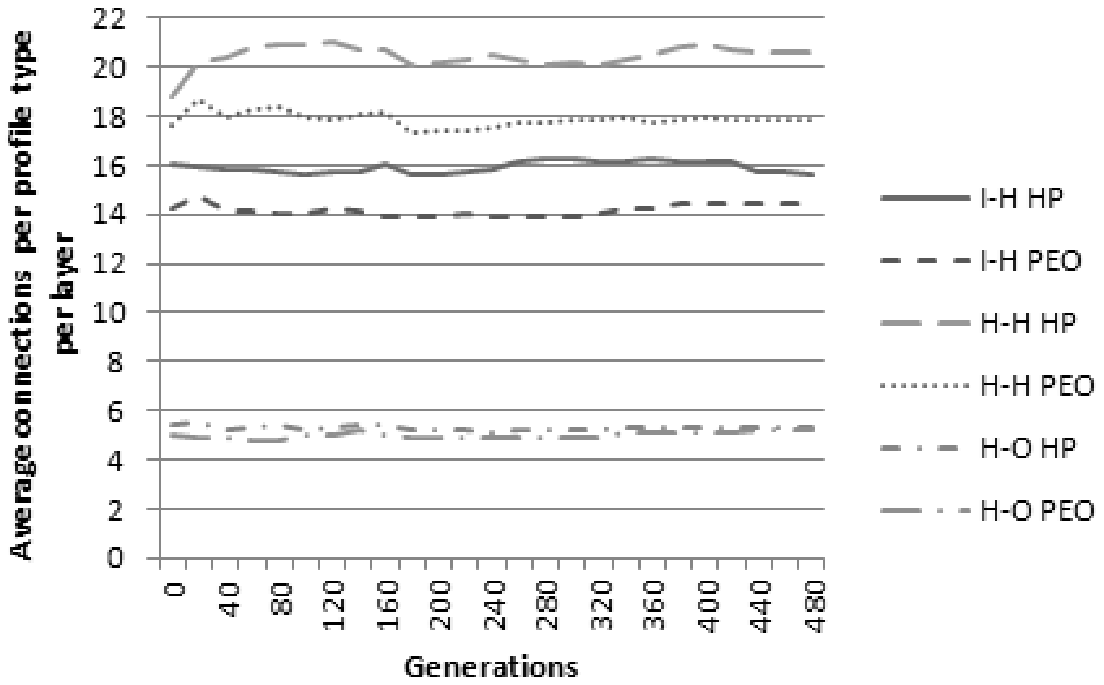,width=6.5cm,height=3.5cm}}\\
\end{center}
\caption[]{Number of connections belonging to (a) each profile type overall in MEM networks (b) each profile type per layer in MEM networks. M = mean of all networks, B = numbers from best network of each experimental repeat.}
\label{dyn-regimeperlayer}
\end{figure}

\subsection{Analysis of Synapse Variability}

\subsubsection{Variable Memristor Networks}
To permit statistical tests on the $\beta$ parameter to correctly express its effect on synapse performance, the $\beta$ values of both synapse types were combined into a single scale.  The total range of $\beta$, including both synapse types, was 199.  HP-governed synapses ({\em type}=1) had possible $\beta$ values between 1 and 101; any $\beta >$ 101 was considered a PEO-PANI-governed synapse with {\em type}=2, and $\beta$ in the range 1-100.  A PEO-PANI-governed profile with $\beta$=50 would therefore have a recalculated value of 101+50=151.

Variation appears in the distribution of profile types in MEM networks, shown in  Fig.~\ref{dyn-regimeperlayer}(a).  Synapses are more frequently governed by HP profiles (average 41.4 synapses per network) than PEO-PANI-governed synapses (average 36.9 synapses per network, p$<$0.05).  More prolific use of HP-governed profiles, which provide lower average efficacy, can be seen as a way to balance network activity given the increased connectivity of MEM networks in general.  Fig.~\ref{dyn-regimeperlayer}(b) shows the distribution of profile types per layer through time. It was observed that HP-governed profiles were more frequently found connecting two hidden-layer neurons (20.6 vs. 17.8 average PEO-PANI-governed connections, p$<$0.05), but were less frequent than PEO-PANI profiles when connecting sensors (input layer) to the hidden layer.  This shows that values of $\beta$ are selected to permit swifter weight change per STDP event when connecting sensors into the network, enabling a more expedient network response to changing environmental conditions (mainly to allow for different actions to be more quickly calculated from similar environmental conditions when R1 changes to R2).

\subsubsection{Variable RSM Networks} 
No differences were found with respect to the type (inhibitory/excitatory) of neuron that synapses of each $S_n$ connect --- p-values range from 0.59 to 0.99.  We note that lower $S_n$ ($S_n$=2-3) synapses are found connecting input neurons to the hidden layer, with $S_n$=4-6 synapses more prevalent when connecting two hidden-layer neurons, suggesting that fast-switching synapses are required to immediately generate activity within the network.  IR sensors have lower $S_n$ than light sensors as they trigger only when near obstacles and so must be able to quickly switch to peturb network output and avoid the obstacle.  

No significant results were observed with respect to the type of neuron (excitatory or inhibitory) the synapses were presynaptic/postsynaptic to in RSM networks (all p $>$0.05), reinforcing the notion that the timing of weight switches within the networks requires subnetworks of synapses with varying temporal behaviour.  This is opposed to MEM networks which assign distinct roles to individual synapses based on their behaviour under STDP.    

\subsubsection{STDP} The ability of a network to generate high quality overall solutions despite the in-trial movement of the goal state is implicitly linked to its ability to alter its internal dynamics during runtime using STDP.  The need to search this additional space during GA application is offset by the increased power of the synapses (for MEM networks) or the increased power of the possible internal network dynamics (for RSM networks). For MEM and RSM networks, the ability of a variable synapse to tailor its behaviour more accurately than the other synapse types is recognised and harnessed by the evolutionary process, resulting in the observed differences in best fitness and average fitness.  

\begin{figure}[ht!]
\begin{center}

\subfloat[]{ \epsfig{file=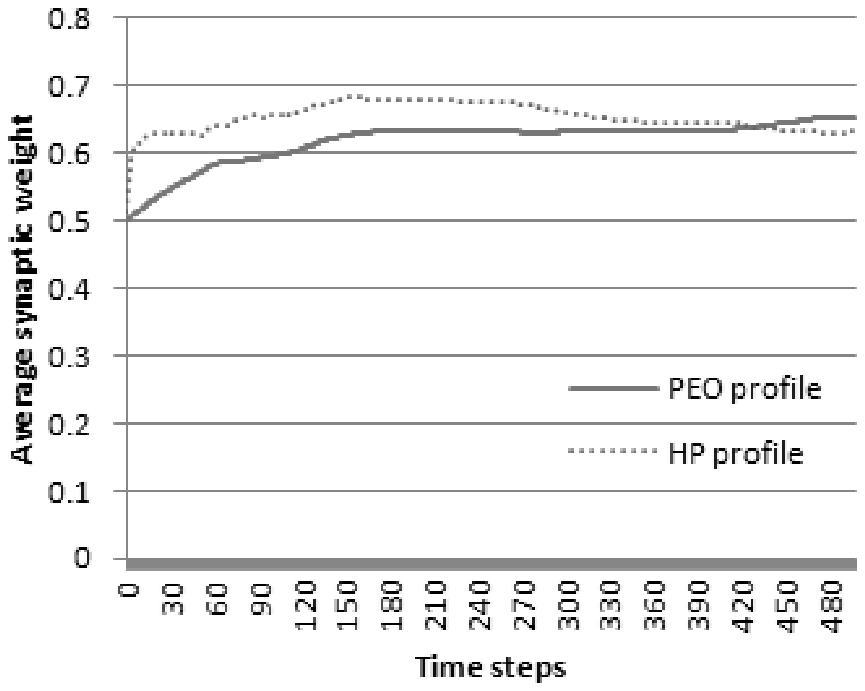,width=4.4cm ,height=3.5cm}}
\subfloat[]{ \epsfig{file=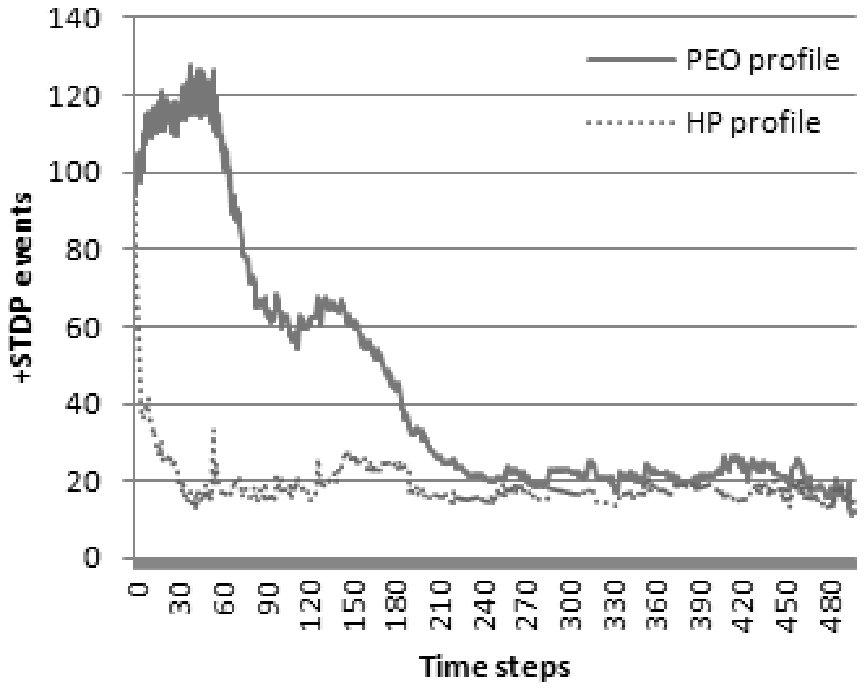,width=4.4cm,height=3.5cm}}
\subfloat[]{ \epsfig{file=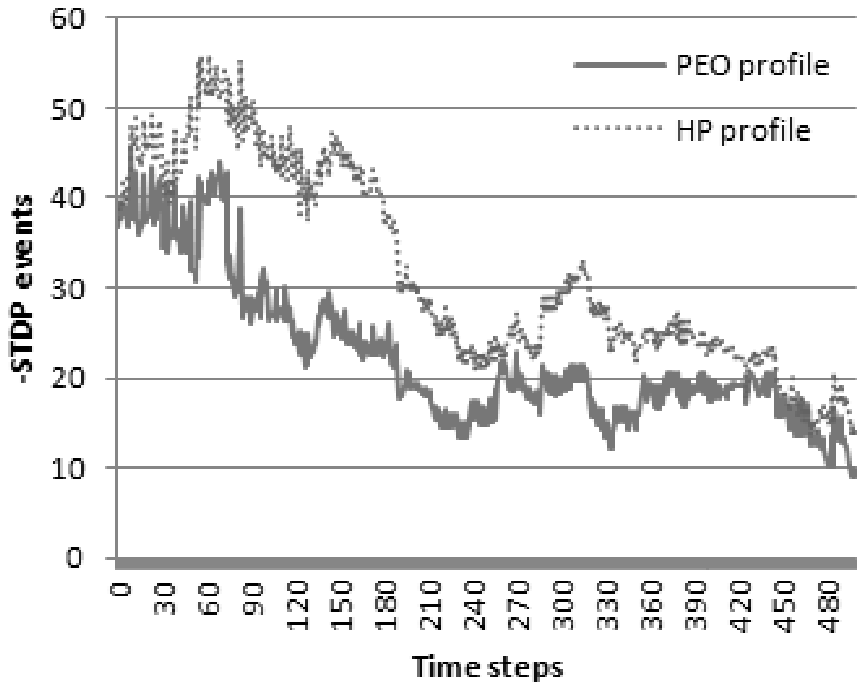,width=4.4cm,height=3.5cm}}\\

\end{center}
\caption[]{(a) average synaptic weight (b) average positive STDP (c) average negative STDP per profile type for the MEM networks in the T-maze.}
\label{dyn-nonlin-stdp}
\end{figure}	

\begin{figure}[ht!]
\begin{center}

\subfloat[]{ \epsfig{file=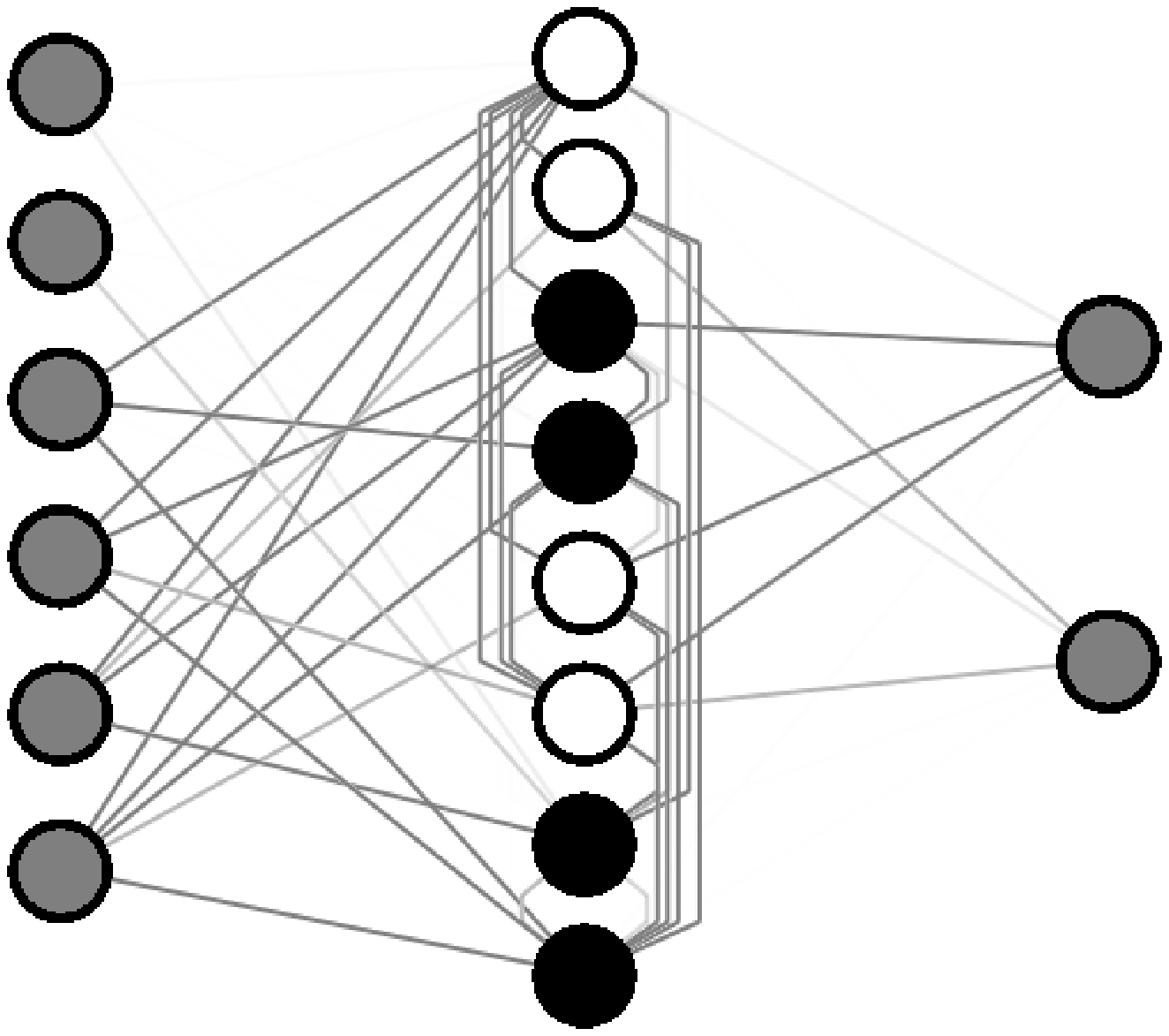,width=4.4cm ,height=3.5cm}}
\subfloat[]{ \epsfig{file=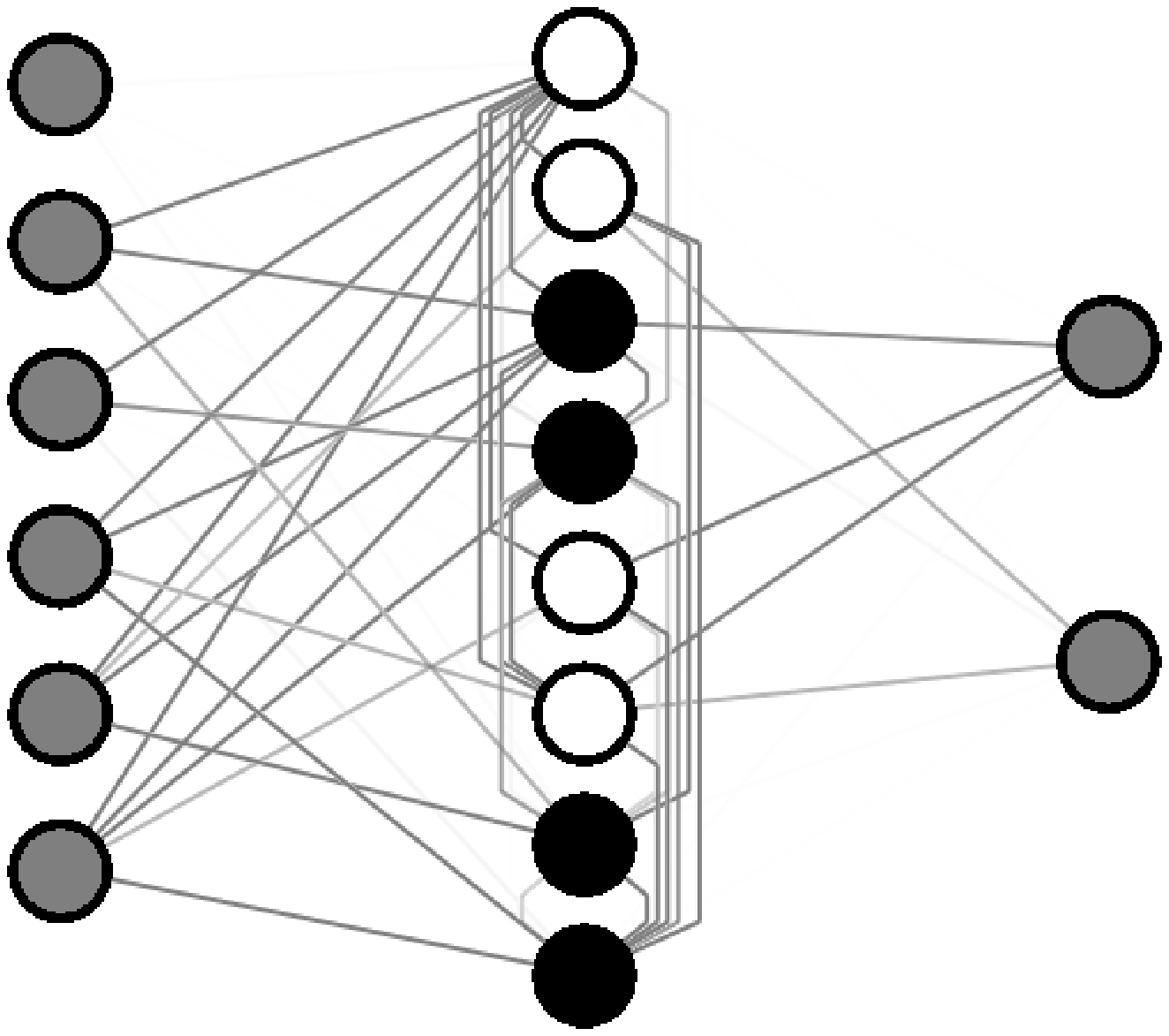,width=4.4cm,height=3.5cm}}
\subfloat[]{ \epsfig{file=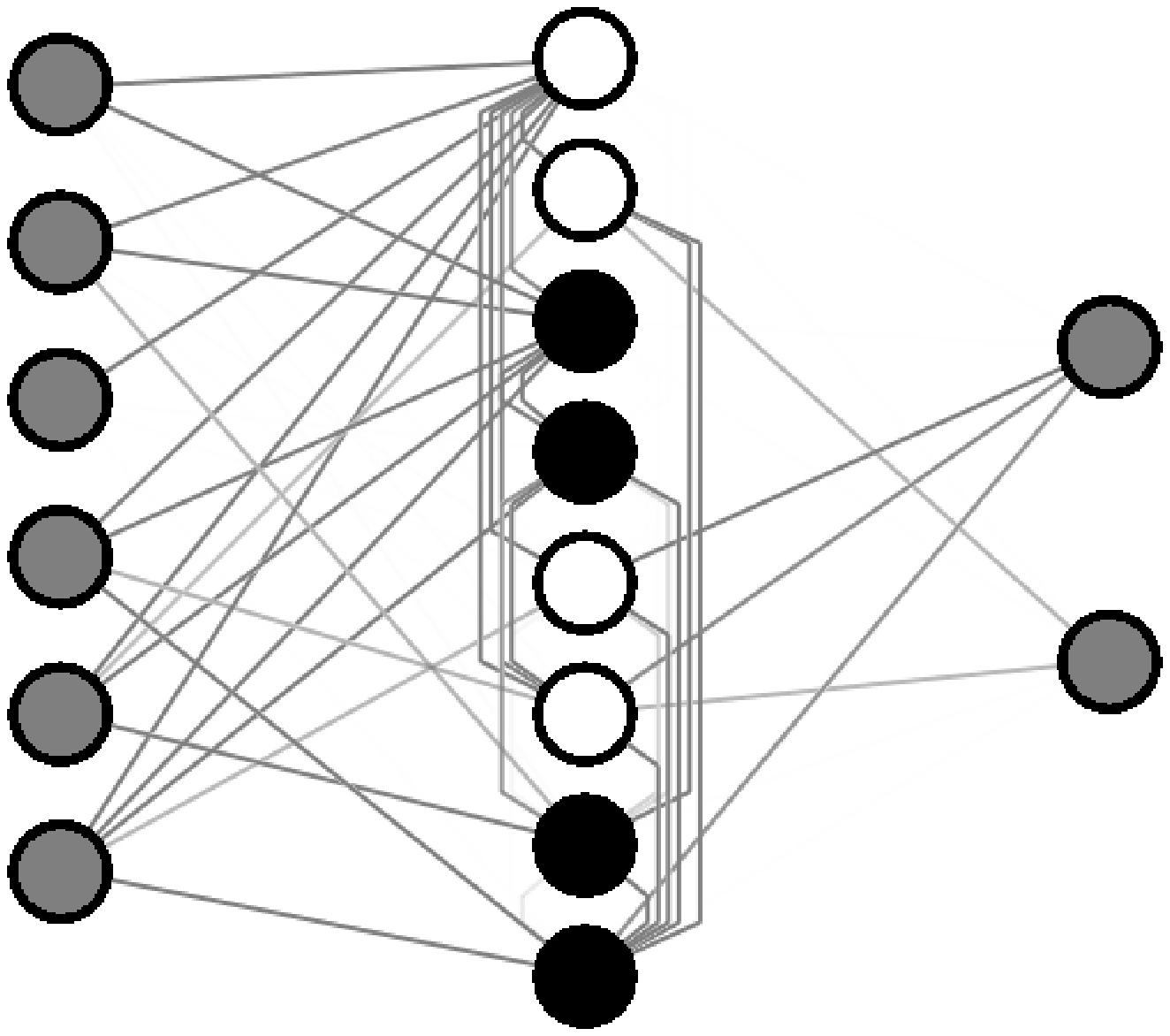,width=4.4cm,height=3.5cm}}\\

\end{center}
\caption[]{Synapse strengths(a)  after the first 50 timesteps (b) after finding the first reward zone (c) after adapting to find the second reward zone for the best MEM network in the T-maze.}
\label{mem-stdp}
\end{figure}	

Use of STDP by the best  MEM network is shown in Fig.~\ref{dyn-nonlin-stdp}, and to give an example of the network's in-trial variance, the network itself is provided in Fig.~\ref{mem-stdp}.  HP-governed profiles were found to quickly reduce synaptic efficacy to the left (right) motor, causing peturbation of calculated action during turn by bringing that motor below the ``high activited'' threshold.  PEO-PANI-governed MEM profiles to the same motor were used to swiftly increase the level of spiking activity (usually in response to a light sensor surpassing/coming under some threshold) until a ``forwards'' action was calculated after the turn was completed.  HP profiles were statistically (p$<$0.05) more likely to be found reducing a synapse's efficacy.  In contrast, PEO-PANI-governed profiles were statistically  (p$<$0.05) more likely to be found increasing a synapse's efficacy.   PEO-PANI-governed profiles experience statistically  (p$<$0.05) more positive STDP and statistically  (p$<$0.05) fewer negative STDP events than HP-governed profiles.  These findings seem to concur with previous work by~\cite{howardTEC}, in which static PEO memristors were found to be more suited to conducting weight through the network (with the opposite being true for HP memristors).  The evolutionary process harnessed the differing profiles by placing PEO memristors where they would receive the most positive STDP.  

\begin{figure}[ht!]
\begin{center}

\subfloat[]{ \epsfig{file=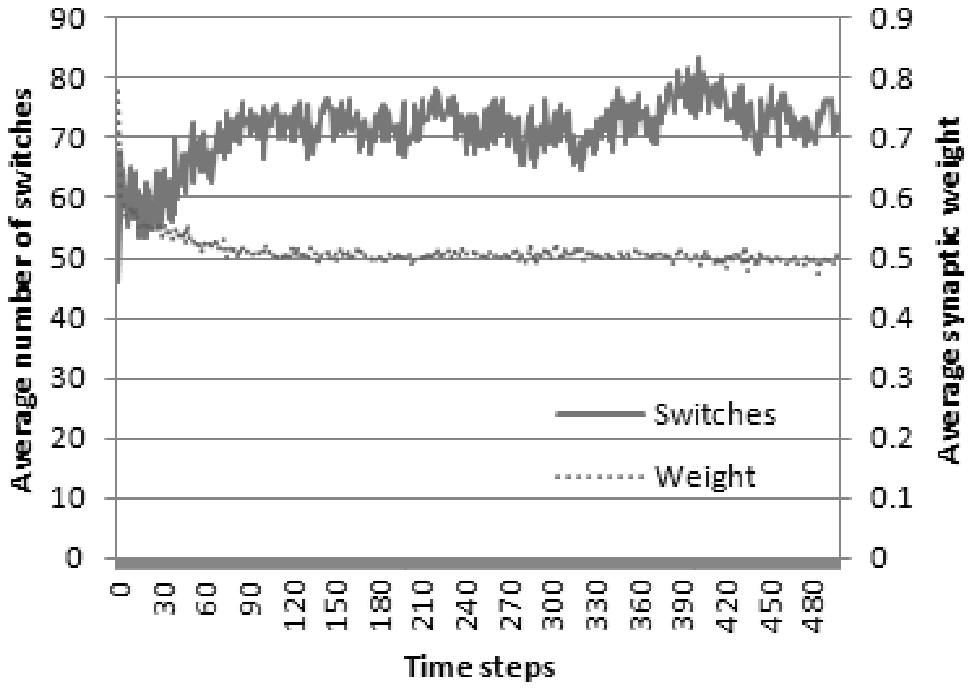,width=7cm ,height=3.5cm}}
\subfloat[]{ \epsfig{file=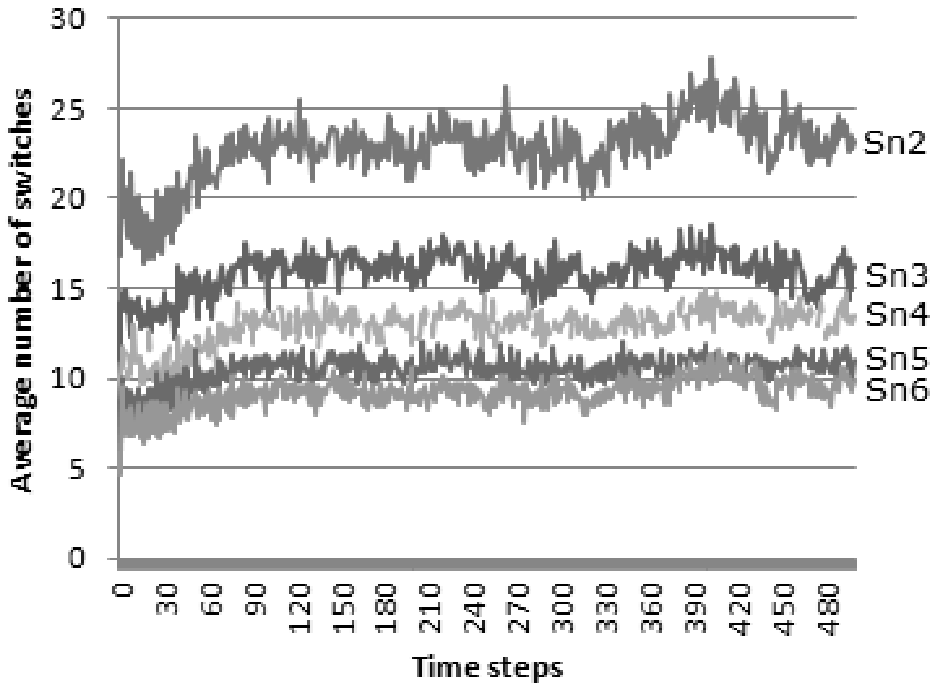,width=7cm,height=3.5cm}}\\

\end{center}
\caption[]{(a) average switch frequency (left axis) / average synaptic weight (right axis) (b) average switch frequency per $S_{n}$ for the RSM networks in the T-maze.}
\label{dyn-lin-stdp}
\end{figure}

RSM networks generated highly fit behaviour via the ability to rapidly vary the network dynamics in three main ways; (i) to perform additional ``connection selection'' in-trial e.g. to switch a synapse to a given state and leave it there; (ii) as (i) but varying the connectivity map of the network multiple times based on the sensory input; (iii) in the creation of weight oscillators in the network, whereby the firing on the neurons and switching of the synapses synchronised through time to generate appropriate output actions from a subgroup of neurons.  In the third case, the input state was found to perturb both the firing pattern of the neurons and weight-switching pattern of the synapses to generate e.g. turning actions when required.  All other network types using STDP relied on numerous repeated events of a particular polarity to provide large increases in efficacy, whereas the RSM could switch back and forth multiple times in a short number of time steps --- more expedient binary switching allows for output to be more quickly altered for a given input configuration.  RSM networks experience a gradual increase in STDP throughout the lifetime of the network.  Figs.~\ref{dyn-lin-stdp}(a) and (b) show the behaviour of the best RSM network from each run through the first 500 timesteps.  The switching profile itself shows two peaks of activity, at time steps 90 and 400, which correspond to the approximate turning times to reach R1 and R2 respectively.   The contribution to switching frequency per $S_n$ is shown in Fig.~\ref{dyn-lin-stdp}(b) --- we observe significant disparity between all $S_n$ types in this regard (all p$<$0.05).  Lower $S_n$ synapses represent more variable STDP profiles as they have higher maximum switching frequencies.  

Fewer overall STDP events occur in RSM networks than MEM networks, presumably because (i) consecutive STDP events are more difficult to attain (ii) each switch can have a more dramatic effect on the activity of the network.  These results suggest that the casting of synapses into roles is only possible when using memristors as synapses, as RSM synapses do not display these relative disparities between $S_n$ types, or a sensitivity to incoming voltage polarity.

\section{Conclusions}

In this study we have introduced the notion of a variable RM and analysed its synaptic performance when compared to  static Resistive Memories and benchmark connections in a dynamic robotics scenario.  Our hypothesis was that the additional degrees of functional freedom afforded to the variable Resistive Memories allowed them to outperform these other synapses in key areas.  Experimental findings supported this hypothesis, as variable Resistive Memories of both kinds generated higher quality solutions than the other synapse types.

Results suggest that self-adapation of the characteristic resistance profile of both variable Resistive Memories is harnessed by the evolutionary process to provide variable plastic networks with more implicit degrees of freedom than the other network types.  Importantly, the need to explore additional search space (especially in the case of $\beta$) was found to be non-disruptive (and in most cases beneficial) with respect to network performance, whilst providing a more flexible synaptic representation.  

The inclusion of self-adaptive mutation parameters with a neuro-evolutionary approach is likely to be necessary for the autonomous emergence of Neuromorphic processing units.  This study presents a candidate implementation that allows for the formation of such task-specific neural groupings.  Futhermore, our results enforce the view that this kind of approach may be used to guide the synthesis requirements of functional memristor/RSM hardware systems.  Trials on different task types may provide insights into the optimal composition of such systems on a per-task basis.

The main benefit of Resistive Memory STDP over other STDP implementations lies in hardware realisation, as the efficacy (and, in the case of the memristor, activity) of the synapse is stored in the nonvolatile physical state of the device and thus does not require simulation.  Possible future research directions include hardware and mixed-media implementations, provided the two Resistive Memory types can be integrated into the same circuit architecture.  It is postulated that Resistive Switching Memories would be easier to implement in hardware due to their mechanically simpler discrete switching behaviours, which would require less finely-tuned manufacturing and be more tolerant of errors during this process.  For synapses implementing approximately the same behaviour, it is likely that the same materials could be used, with finely controlled variations during synthesis required to achieve the desired behaviours.  Where synapses require radically different functionality, mixing of heterogeneous materials may be required --- considering the recent discovery of myriad memristors with varied behaviour at similar scales, it is likely that any evolved network will have a compatible behavioural hardware equivalent.  We note that titanium dioxide additionally allows for memristive behaviour and binary switching to be elicited from the same material.  As well as providing more functional degrees of freedom to the synapse,  evolution could potentially control switching between the behaviours to autonomously create task-optimal Neuromorphic subarchitectures, as well as online synaptic transformations via targetted irradiation for self-repair or self-modification.

\bibliographystyle{apalike}
\bibliography{varmem}  

\end{document}